# Dual-constrained Deep Semi-Supervised Coupled Factorization Network with Enriched Prior

Yan Zhang[1], Zhao Zhang[2,3,4], Yang Wang[2,3,4], Zheng Zhang[5,6], Li Zhang[1], Shuicheng Yan[7], *Fellow, IEEE*, and Meng Wang[2,3,4], *Fellow, IEEE*

[1] School of Computer Science and Technology, Soochow University, Suzhou, China
[2] School of Computer Science and Information Engineering, Hefei University of Technology, Hefei, China
[3] Key Laboratory of Knowledge Engineering with Big Data (Ministry of Education), Hefei University of Technology, China
[4] Intelligent Interconnected Systems Laboratory of Anhui Province, Hefei University of Technology, Hefei, China
[5] Bio-Computing Research Center, Harbin Institute of Technology, Shenzhen, China
[6] Pengcheng Laboratory, Shenzhen, China
[7] Sea AI Labs, Singapore

*Abstract*— **Nonnegative matrix factorization is usually powerful for learning the parts-based "shallow" representation, however it fails to discover deep hidden information within both the basis concept and representation spaces. In this paper, we therefore propose a new Dual-constrained Deep Semi-Supervised Coupled Factorization Network (DS$^2$CF-Net) for learning hierarchical representations. DS$^2$CF-Net is formulated as the joint partial-label and structure-constrained deep factorization network using multi-layers of linear transformations, which coupled updates the basis concepts and new representations in each layer. An error correction mechanism with feature fusion strategy is also integrated between consecutive layers to improve the representation ability of features. To improve the discriminating abilities of both representation and coefficients in feature space, we clearly consider how to enrich the prior knowledge by the coefficients-based label prediction, and incorporate the enriched prior knowledge as the additional label and structure constraints. To be specific, the label constraint enables the intra-class samples to have the same coordinate in the feature space, while the structure constraint forces the coefficients in each layer to be block-diagonal so that the enriched prior knowledge are more accurate. Besides, we integrate the adaptive dual-graph learning to retain the locality structures of both the data manifold and feature manifold in each layer. Finally, a fine-tuning process is performed to refine the structure-constrained matrix and data weight matrix in each layer using the predicted labels for more accurate representations. Extensive simulations on public databases show that our method can obtain state-of-the-art performance.**

***Keywords****— Deep semi-supervised coupled factorization network; representation learning; dual constraints; clustering; enriched prior; error correction; fine-tuning of features*

## I. Introduction

For high-dimensional data analysis in emerging computer vision applications and, one core problem is how to obtain the compact expression with strong representation ability from complex and high-dimensional data [55-58]. To compute strong and effective representations, different algorithms can be used, among which Matrix Factorization (MF) is one of the widely-used techniques for representation [1-5][48-52]. Classical MF methods include Singular Value Decomposition (SVD) [2], Vector Quantization (VQ) [3], Nonnegative Matrix Factorization (NMF) [4] and Concept Factorization (CF) [5], etc. It is noteworthy that NMF and CF use the nonnegative constraints on the factorization matrices, which enables them to obtain parts-based representations that correspond to the useful distinguishing features for subsequent clustering and classification [4-5]. Specifically, NMF and CF aim at decomposing a given data matrix $X$ into two or three nonnegative matrix factors whose product is the approximation to $X$ [4-5], where one factor contains the basis vectors capturing the high-level features and each sample is reconstructed by a linear combination of the basis vectors. The other nonnegative factor corresponds to the learnt new representation.

CF offers an obvious advantage over NMF, i.e., it can be performed in kernel space and any other representation space, however they both cannot encode the local geometry of features and also fail to apply the label information even if available. To handle the locality preserving issue, some graph regularized methods have been proposed, e.g., Graph Regularized NMF (GNMF) [6], Graph-Regularized LCF (GRLCF) [8], Locally Consistent CF (LCCF) [7], Graph-Regularized CF with Local Coordinate (LGCF) [9], Dual Regularization NMF (DNMF) [12] and Dual-graph regularized CF (GCF) [13]. These algorithms usually use the graph Laplacian to smooth the representation and encode the geometry information of the data space. Different from GNMF and LCCF, both DNMF and GCF can not only retain the geometrical structures of the data manifold but also the feature manifold using the dual-graph regularization strategy [10-13]. Although the above algorithms have obtained encouraging clustering results by considering the locality properties, they still suffer from some shortcomings: 1) high sensitivity and tricky optimal determination of the number $k$ of nearest neighbors [14-15]; 2) separating the graph construction from the matrix factorization by two independent steps cannot ensure the pre-encoded weights to be optimal for subsequent representation; 3) they cannot take advantage of the label information to improve the representation and clustering due to the unsupervised nature, similarly as NMF and CF. To improve the discriminating ability of the MF, some semi-supervised algorithms have been proposed, such as Constrained Nonnegative Matrix Factorization (CNMF) [16], Constrained Concept Factorization (CCF) [17] and Semi-supervised GNMF (SemiGNMF) [6]. SemiGNMF incorporates partial label information into the graph construction, while CNMF and CCF obtain the representations consistent with the known label information by defining an explicit label constraint matrix, so that the original labeled samples sharing the same label can be mapped into the same class in the feature space. Although CNMF, SemiGNMF and CCF can use the label information of labeled samples clearly, they still fail to fully utilize the unlabeled samples. Since they did not consider predicting the labels of the unlabeled samples and mapping them into their respective subspaces in the feature space as well by learning an explicit label indicator matrix for the unlabeled data. In addition, CNMF, SemiGNMF and CCF also cannot self-express the input data in a recovered feature space. Although preserving the local geometrical structures and incorporating the supervised prior information can improve the representation abilities of NMF and CF effectively, however all aforementioned MF-based models still suffer from a common drawback, i.e., they use a single-layer mode so that they can only discover the "shallow" features while cannot discover deep hidden features and hierarchical information that have been proved to be important and useful for representation learning.

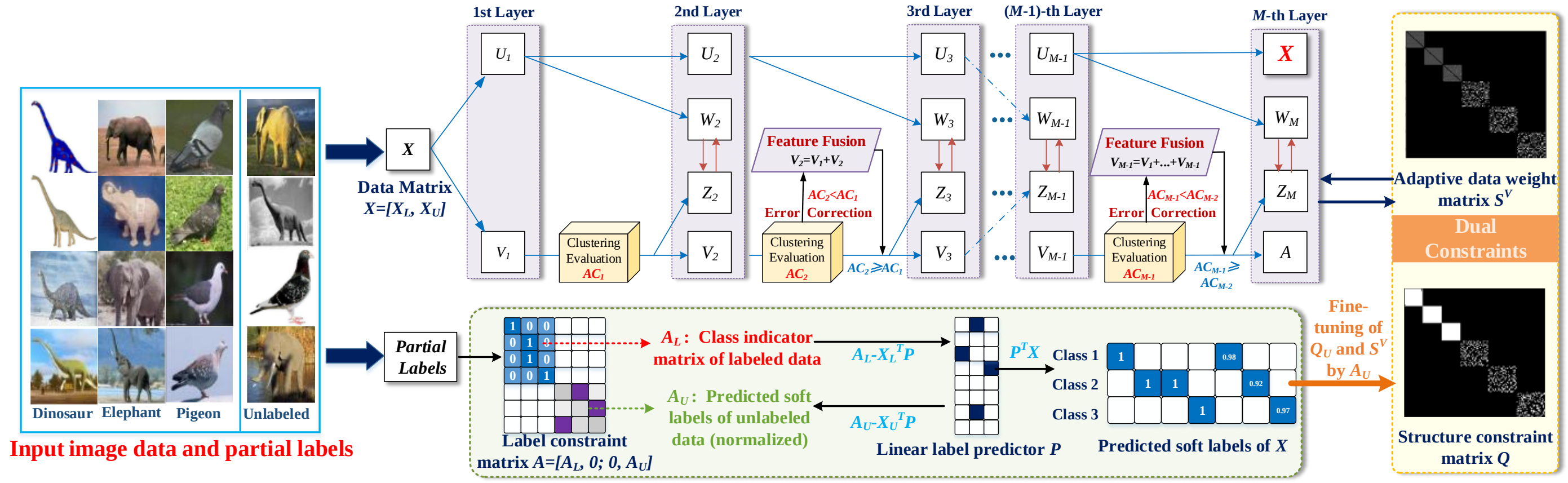

**Figure 1:** The flowchart and learning principle of our proposed DS$^2$CF-Net framework.

In this paper, we propose a novel deep semi-supervised self-expressive coupled MF strategy that can represent the input data more appropriately using partially labeled data and a deep structure. The main contributions of this work are shown below:

(1) Technically, a novel supervised prior enrichment guided Dual-constrained Deep Semi-Supervised Coupled Factorization Network (shortly, DS$^2$CF-Net) is proposed. To learn the hierarchical coupled representation and extract hidden deep features, we seamlessly integrate the deep coupled semi-supervised concept factorization, prior knowledge enrichment, self-expressive discriminating representation, and the joint label/structure constraints into a unified framework. To encode the deep features accurately, we design a novel updating strategy for the deep MF, i.e., it coupled optimizes the basis vectors and representation matrix in each layer by learning with partial labeled data. Fig.1 illustrates the flowchart of our DS$^2$CF-Net.

(2) For discriminating representation, the innovations of our DS$^2$CF-Net are twofold; 1) it clearly considers enriching the supervised prior knowledge by the joint coefficients-based label prediction; 2) it incorporates the enriched label information as the additional dual label and structure constraints. To enrich the prior knowledge, DS$^2$CF-Net tries to make full use of the unlabeled data by propagating and predicting the labels of unlabeled data using a robust label predictor learned from the labeled data. The dual-constraints are included to improve the discriminating power of the learned representations. Specifically, the enriched prior knowledge based label constraint can enable the originally labeled samples of one class and the unlabeled data with the predicted same label to have the same coordinate in feature space. The enriched prior knowledge based structure constraint forces the self-expressive coefficients matrix to be block-diagonal in each layer so that the manifold is more smooth and accurate for label prediction. Besides, to make full use of the predicted labels, we also consider refining the structure constraint matrix and the data weight matrix to further make the learned representations better using the obtained soft labels in each layer.

(3) To obtain neighborhood-preserving higher-level representations, DS$^2$CF-Net presents a self-weighted dual-graph learning strategy in each layer, i.e., optimizing the graph weights jointly with the MF. Specifically, in each layer, DS$^2$CF-Net performs an adaptive weight learning over the deep basis vector graph and deep feature graph through minimizing the reconstruction errors based on the deep basis vectors and deep features at the same time. The self-weighted dual-graph learning can also avoid the tricky issue of selecting the optimal number of the nearest neighbors, which is suffered in most existing locality preserving MF models. More importantly, such operation can enable the model to obtain adaptive neighborhood preserving deep basis vectors and deep features for enhancing the representations.

(4) When the number of layers increases, to obtain more stable and reliable representation and clustering results, we incorporate an error analysis mechanism and a feature fusion strategy into our framework. Specifically, when the clustering result on the features obtained in the current layer is lower than that of the previous layer, a feature fusion will be operated. Note that such operation can effectively prevent the issue that the performance decreases fast with the increase of layers.

We outline the paper as follows. Section II briefly reviews the related work. We present DS$^2$CF-Net in Section III. In Section IV, we show the optimization procedures of our DS$^2$CF-Net. Section V describes the simulation settings and results. Finally, the paper is concluded in Section VI.

## II. Related Work

In this section, we introduce the related single-layer and multi-layer frameworks to our proposed DS$^2$CF-Net.

### A. *Realted Single-layer CF based Frameworks*

We first show the closely-related single-layer CF and its variants.

**Concept Factorization (CF) [5].** Given a nonnegative data

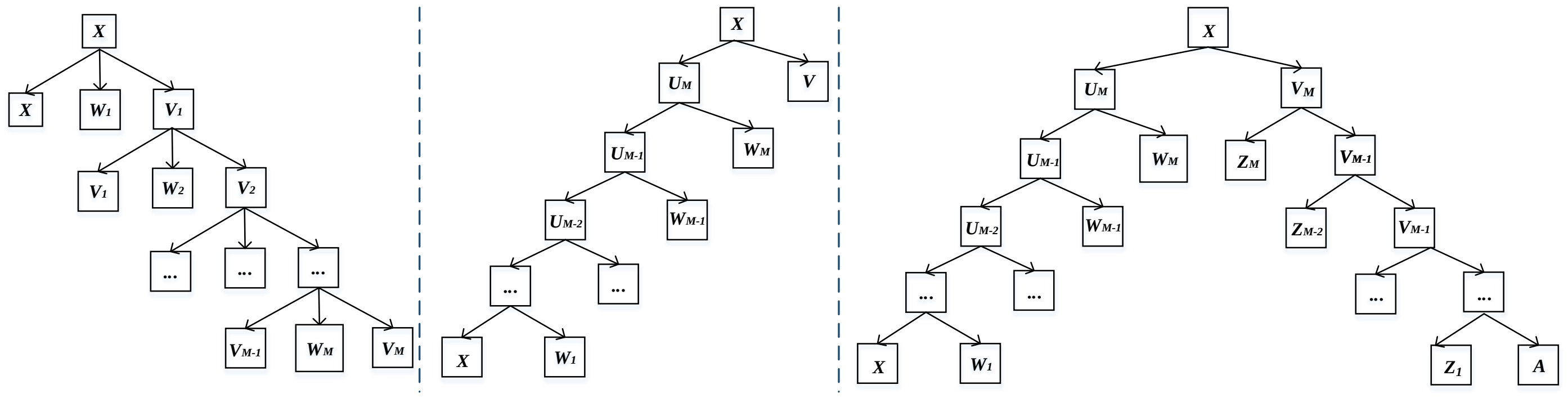

(a) (b) (c)

**Figure 2:** Architecture comparison of existing multilayer MF framework, including (a) Traditional multilayer CF model (e.g., MNMF, MCF and GMCF); (b) DSCF-Net model; (c) Our proposed DS[2]CF-Net.

matrix $X=[x_1,x_2,...,x_N]\in\mathbb{R}^{D\times N}$, where $x_i$ is a sample vector, $N$ is the number of samples and $D$ denotes the original dimension of the input space. Denote by $U\in\mathbb{R}^{D\times r}$ and $V\in\mathbb{R}^{N\times r}$ two nonnegative matrices whose product $UV^T\in\mathbb{R}^{D\times N}$ denotes the approximation to $X$, where $r$ is the rank. By representing each basis by a linear combination of $x_i$, i.e., $\sum_{i=1}^{N} w_{ij}x_i$, where $w_{ij}\geq 0$, then CF proposes to solve the following minimization problem:

$$O=\left\|X-XWV^T\right\|_F^2,\ \ s.t.\ W,V\geq 0\ , \tag{1}$$

where $W=[w_{ij}]\in\mathbb{R}^{N\times r}$, $XW$ approximates the bases, $V^T$ is the learned representation of $X$, which can be applied for clustering, and $V^T$ is the transpose of the representation matrix $V$.

**Self-Representative Manifold CF (SRMCF) [24].** SRMCF integrates the adaptive neighbor structure and manifold regularizer into the CF framework. Specifically, it considers $WV^T$ in CF as the coefficient matrix based on the dictionary of the raw data matrix $X$. Then, it incorporates the self-representation with the adaptive neighbor structure to assign neighbors for all samples. The objective function of SRMCF is defined as

$$\begin{aligned}O=&\left\|X-XWV^T\right\|_2^2+\lambda_1\sum\nolimits_{i,j=1}^{N}\left\{\left\|\left(WV^T\right)_i-\left(WV^T\right)_j\right\|_2^2\Theta_{ij}+\xi\Theta_{ij}^2\right\}\\&+\lambda_2 tr\left(V^TL^VV\right),\ \ s.t.\ W\geq 0,V\geq 0,\ \forall i\ \Theta_i^T\mathbf{1}=1,\ 0\leq\Theta_i\leq 1\end{aligned}, \tag{2}$$

where $\mathbf{1}$ is an all-ones column vector, $\xi$ is a positive trade-off parameter, and $\Theta_{ij}$ denotes the probability of $x_j\in[x_1,x_2,...,x_N]$ (excluding itself) being connected to $x_i$ as a neighbor. $\Theta_i\in\mathbb{R}^{N\times 1}$ is a vector with the $j$-th element as $\Theta_{ij}$. Note that the constraints $\Theta_i^T\mathbf{1}=1$ and $0\leq\Theta_i\leq 1$ are used to ensure the probability property of $\Theta_i$. $L^\Theta$ is the Laplacian matrix of $\Theta$ and $L^V$ is a predefined Laplacian matrix by 0-1 weight based on the Euclidean distances between each sample $x_j\in[x_1,x_2,...,x_N]$ as [47]. $\lambda_1$ and $\lambda_2$ are two parameters. Note that SRMCF still suffers from the tough choice of the number of nearest neighbors of each sample.

**Dual-graph regularized CF (GCF) [13].** GCF introduces the graph regularizers of both the data manifold and feature manifold into CF simultaneously by constructing a $k$ nearest neighbor data graph $G^V$ and a $k$ nearest feature graph $G^U$. Then, GCF uses the 0-1 weighting scheme for $G^V$ and $G^U$ and defines the corresponding weight matrices $S^V$ and $S^U$ as follows:

$$\begin{aligned}\left(S^V\right)_{js}&=\begin{cases}1 & if\ x_s\in N_k\left(x_j\right)\\ 0 & otherwise\end{cases};\ j,s=1,2...,N\\ \left(S^U\right)_{js}&=\begin{cases}1 & if\ x_s^T\in N_k\left(x_j^T\right)\\ 0 & otherwise\end{cases};\ j,s=1,2...,M\end{aligned}, \tag{3}$$

where $N_k(x_j)$ denotes the set of the $k$ nearest neighbors of $x_j$. The graph Laplacian over $G^V$ and $G^U$ are defined as $L^V=D^V-S^V$ and $L^U=D^U-S^U$, where $D^V$ and $D^U$ are diagonal matrices with entries being $\left(D^V\right)_{jj}=\sum_s\left(S^V\right)_{js}$ and $\left(D^U\right)_{jj}=\sum_s\left(S^U\right)_{js}$. Finally, the objective function of GCF is formulated as

$$O=\left\|X-XWV^T\right\|_F^2+\alpha tr\left(V^TL^VV\right)+\beta tr\left(W^TL^WW\right), \tag{4}$$

where $L^W=X^TL^UX$, $\alpha$ and $\beta$ are parameters. Clearly, GCF has the difficulty issue to choose the optimal $k$ on various datasets.

**Constrained CF (CCF) [17].** To improve the discriminating power, CCF extends CF to the semi-supervised scenario by using label information of the labeled data as an additional constraint. Suppose that the data matrix $X$ contains a labeled sample set $X_L\in\mathbb{R}^{D\times l}$ and an unlabeled set $X_U\in\mathbb{R}^{D\times u}$, that is, $l+u=N$ and $X=[X_L,X_U]\in\mathbb{R}^{D\times(l+u)}$, where $l$ and $u$ are the numbers of labeled and unlabeled samples respectively, then CCF guides the constrained CF by defining a label constraint matrix $A$. Denote by $A_L\in\mathbb{R}^{l\times c}$ the class indicator matrix defined on labeled data, where $c$ is the number of classes. The element $(A_L)_{ij}$ is defined as 1 if $x_i$ is labeled as the $j$-th class, and 0 otherwise. Since CCF did not define an explicit class indicator for $X_U$ and simply used an identity matrix $I_{u\times u}$ of dimension $u\times u$ for unlabeled data. Thus, the overall label constraint matrix $A$ is defined as

$$A=\begin{bmatrix}(A_L)_{l\times c} & 0\\ 0 & I_{u\times u}\end{bmatrix}\in\mathbb{R}^{(l+u)\times(c+u)}. \tag{5}$$

To ensure the data points sharing the same label to be mapped into the same class in low-dimensional space (i.e., same $v_i$), CCF imposes the label constraints by an auxiliary matrix $Z$:

$$V=AZ\ . \tag{6}$$

By substituting $V=AZ$ into CF, CCF finds a non-negative matrix $W\in\mathbb{R}^{N\times r}$ and a non-negative auxiliary matrix $Z\in\mathbb{R}^{(c+u)\times r}$ from the following objective function:

$$O = \left\| X - XWZ^T A^T \right\|_F^2 , \ \ s.t.\ W, Z \geq 0 . \quad (7)$$

## B. Related Deep/Multilayer MF Frameworks

We then introduce the architectures of several related deep/multilayer matrix factorization algorithms.

**Traditional multilayer MF.** The multilayer MF methods of this category usually use the output of the previous layer (i.e., intermediate representation *V*) as the input of subsequent layer directly, without properly considering to optimize the representation and basis vectors in each layer. Classical methods include MNMF, MCF and GMCF, etc. These methods aim to minimize the objective function in each layer independently and simply use $V_{m-1}$ ( $m \geq 2$ ) obtained in the *(m-1)*-th layer as the input of the *M*-th layer. That is, they cannot ensure the intermediate representation to be a good representation for subsequent layers, which may cause the degraded performance. We show the multilayer structure of this category methods in Fig.2(a).

**Optimized deep MF models.** The recent fast developments of deep learning have led to a renewed interest in designing the deep or multi-layer MF [18-23][49-51] for deep representation learning and clustering. One of the most widely-used approach to extend the single-layer model to the deep model of *M*-layers is to iteratively take the outputted representation of the last layer as the inputs of the next layer directly for further MF [18-21], where *M* denotes the number of layers, such as Multilayer NMF (MNMF) [18], Multilayer CF (MCF) [19], Spectral Unmixing using Multilayer NMF (MLNMF) [20] and Graph regularized multilayer CF (GMCF) [21]. However, such a strategy may be invalid and even unreasonable in practice, because in this case the learned representation of the first layer determines the learning abilities of the whole framework, while most existing models cannot ensure this issue. In other words, one cannot ensure that the output of the last layer is already a good representation, so directly feeding it to the next layer may mislead and degrade the learning power of subsequent layers. To address these issues, the other popular and optimized way is to discover hidden deep feature information by adopting multiple layers of linear transformations and updating the basis vectors or feature representations in each layer [22-23], such as Weakly-supervised Deep MF (WDMF) [22], Deep Semi-NMF (DSNMF) [23] and Deep Self-representative Concept Factorization Network (DSCF-Net) [43]. In general, WDMF aims at fixing the basis vectors and optimizes the representations in each layer, while DSCF-Net argues that learning a set of optimal basis vectors will be more important and accurate for reconstructing given data by a linear combination of the bases, which aims at optimizing the basis vectors to update the representation matrix in each layer. It is noted that WDMF mainly focuses on the social image understanding tasks, i.e., tag refinement, tag assignment and image retrieval, and the initial input of WDMF is the tagging matrix *F* rather than the data matrix *X* as other MF models. In addition, DSCF-Net also incorporates the subspace recovery process and adaptive locality-preserving power into a united framework for enhancing the feature representations. Different from WDMF and DSCF-Net, DSNMF is just a two-stage approach, where the strategy in the first stage is the same as traditional MNMF, MCF, MLNMF and GMCF, i.e., directly feeding the learned representation matrix of the last layer into the next layer for further MF, and the second stage refines the representation matrices and basis vectors directly based on the outputs of each layer in the first stage using an independent step. It is clear that the refining step can obtain deep features, but the learned deep features are directly based on the first stage. As such, DSNMF will suffer from the same performance-degrading issue as traditional MNMF, MCF, MLNMF and GMCF. In addition, DSNMF also cannot preserve the manifold structures of samples, especially in an adaptive manner, and it also cannot use supervised prior information for the discriminant data representations. Note that we illustrate the multilayer structure of DSCF-Net for uncovering hidden features in Fig.2(b). For comparison, we also illustrate the deep coupled factorization network of our DS$^2$CF-Net in Fig.2(c), from which we can see that DS$^2$CF-Net coupled optimizes the basis vectors and representation matrix in each layer.

# III. Deep Semi-Supervised Coupled Factorization Network (DS$^2$CF-Net)

We introduce the formulation of DS$^2$CF-Net. Given a partially labeled data matrix $X = [X_L, X_U] \in \mathbb{R}^{D\times(l+u)}$, DS$^2$CF-Net performs the semi-supervised representation-based clustering over the whole dataset. As a classical semi-supervised learning setting, the labeled set contains a small proportion of samples, while the unlabeled set contains a large proportion of samples. The base model of DS$^2$CF-Net is built based on the semi-supervised formulation of CCF, i.e., incorporating a label constraint matrix *A* and approximating the representation matrix *V* with *AZ*. However, to enhance the data representation and clustering abilities, DS$^2$CF-Net designs a hierarchical and coupled factorization framework that has *M* layers. Technically, DS$^2$CF-Net is modeled as the formulation of learning *M* updated pairs of representation matrices and basis vectors $XW_1 \ldots W_M$ , and *M* updated label constraint matrices *A*. That is, the label constraint matrix *A* in our DS$^2$CF-Net is alternately updated and enriched over unlabeled data, instead of fixing it as CCF does.

## A. Factorization Model

Before presenting the factorization model, we first describe the initial optimization problem of DS$^2$CF-Net as follows:

$$\begin{aligned} &O_{DS^2CF-Net} = \left\| X - XW_0 ... W_M \left( Z_0 ... Z_M \right)^T A^T \right\|_F^2 + \alpha J_2 + \beta J_3 + \gamma J_1 \\ &s.t.\ \forall_{i \in \{1,2,...,M\}}\ W_i \geq 0,\ Z_i \geq 0 \end{aligned}, \quad (8)$$

where $XW_0 \ldots W_M$ corresponds to the set of deep basis vectors, $(Z_0 \ldots Z_M)^T A^T$ denotes the learned deep low-dimensional representation, $\left\| X - XW_0 ... W_M \left( Z_0 ... Z_M \right)^T A^T \right\|_F^2$ denotes the deep reconstruction error, $J_1$, $J_2$ and $J_3$ will be shown shortly. $W_0$ and $Z_0$ are included to facilitate the description and optimization, and both are fixed to be the identity matrices. The overall label constraint matrix *A* in our network is defined as follows:

$$A = \begin{bmatrix} A_L & 0 \\ 0 & A_U \end{bmatrix} \in \mathbb{R}^{(l\times u)\times(c+c)},\ A_L \in \mathbb{R}^{l\times c}, A_U \in \mathbb{R}^{u\times c} , \quad (9)$$

where $A_L$ is a class indicator matrix for labeled data, which can be easily defined as [17], i.e., $(A_L)_{i,j}=1$ if sample $x_i$ is labeled as the class $j$, and else 0. However, DS[2]CF-Net also computes an explicit class indicator $A_U$ for the unlabeled data to enrich the supervised prior rather than fixing it to be an identity matrix as CCF, which can well group the representations of both the labeled and unlabeled samples based on the enriched supervised prior guided dual label and structure constraints.

According to the self-expressive properties on the coefficient matrix [24], the reconstruction error can be rewritten as

$$\left\|X - XR_M\right\|_F^2,\ where\ R_M = W_0 ... W_M \left(Z_0 ... Z_M\right)^T A^T, \tag{10}$$

where $R_M$ can be regarded as the meaningful coefficient matrix self-expressing $X$. Then, the proposed multi-layer factorization model can be presented as follows:

$$X \leftarrow U_M V_M^T$$
$$\begin{matrix} U_M = U_{M-1} W_M \\ \vdots \\ U_2 = U_1 W_2 \\ U_1 = XW_1 \end{matrix} \quad \text{and} \quad \begin{matrix} V_M = V_{M-1} Z_M \\ \vdots \\ V_2 = V_1 Z_2 \\ V_1 = AZ_1 \end{matrix}, \tag{11}$$

where $U_m$ ($m$=1, 2,.., $M$) is the set of basis vectors of the $m$-th layer, $V_m^T$ ($m$=1, 2,.., $M$) is the low-dimensional representation, $W_m$ ($m$=1, 2,.., $M$) is the intermediate matrix for updating basis vectors and $Z_m$ ($m$=1, 2,.., $M$) is the intermediate auxiliary matrix for updating the representations. It is noteworthy that the factorization model of DS[2]CF-Net does not need to initialize the network using the traditional multi-layer model as DSNMF that initializes the network by directly feeding the learnt representation matrix of the last layer into the next layer for MF. And the deep factorization process of DSNMF completely depends on the intermediate outputs of traditional multi-layer model.

### *B. Enriched Prior based Dual-constraints*

We first describe how to enrich the supervised prior information. DS[2]CF-Net learns a robust label predictor $P \in \mathbb{R}^{D\times c}$ over labeled data by minimizing a label fitness error $\left\|A_L - X_L^T P\right\|_F^2$, where $c$ is the number of classes, which can map each sample $x_i$ into a label space in terms of $P^T x_i$. In addition, DS[2]CF-Net also considers preserving the neighborhood information of the embedded soft labels $P^T X_i$ in the projective label space by self-expressing it with the coefficient matrix $R_M$. The formulation of learning the label predictor $P$ can then be defined as follows:

$$\begin{aligned} J_1 &= \left\|A_L - X_L^T P\right\|_F^2 + \left\|P^T X - P^T X R_M\right\|_F^2 + \left\|P\right\|_{2,1} \\ &= \left\|A_L - X_L^T P\right\|_F^2 + \left\|P^T X - P^T X\left(W_0 ... W_M \left(Z_0 ... Z_M\right)^T A^T\right)\right\|_F^2 + \left\|P\right\|_{2,1} \end{aligned}, \tag{12}$$

where $L_{2,1}$-norm based regularization can enable the label predictor to be robust against the outliers and error in data. In addition, $L_{2,1}$-norm can enable the discriminative labels to be predicted and estimated in a latent sparse feature subspace.

**Enriched prior based label constraint.** After the label predictor $P$ is obtained, we can easily predict the soft label of each unlabeled sample $x_i \in X_U$ as $x_i^T P$. Then, we obtain $A_U$ by using the normalized soft labels that are described as follows:

$$\left(A_U\right)_{ij} = \left(X_U^T P\right)_{ij} \Big/ \sum\nolimits_{j=1}^{c} \left(X_U^T P\right)_{ij}. \tag{13}$$

That is, the normalized soft labels meet the column-sum-to-one constraint $A_U\mathbf{1}=\mathbf{1}$, where $\mathbf{1} \in \mathbb{R}^{c\times 1}$ is a column vector of ones. Note that one recent related work is called Robust Semi-Supervised Adaptive Concept Factorization (RS[2]ACF) [42] has also discussed the partially labeled CF model and considered learning a class indicator $A_U$ for unlabeled data, but our DS[2]CF-Net is different from it in three aspects. First, DS[2]CF-Net is a deep MF model, while RS[2]ACF is a single-layer model. Second, the manifold smoothness for label prediction in DS[2]CF-Net is defined based on the self-expressive deep coefficient matrix in each layer, while RS[2]ACF encodes the manifold smoothness by learning an extra weight matrix and is performed in a single-layer mode. Third, DS[2]CF-Net defines the class indicator $A_U$ based on the normalized soft labels of unlabeled data rather than directly embedding $X_U$ into $P$. Since the predicted soft label value $(A_U)_{ij}$ indicates the probability of each $x_i$ belonging to the class $j$, forcing $A_U\mathbf{1}=\mathbf{1}$ may be more accurate and reasonable.

**Enriched prior based structure constraint.** Since the coefficients $W_0 ... W_M(Z_0 ... Z_M)^T A^T$ can characterize the locality of the features, it should have a good block-diagonal structure, where each block corresponds to a subspace or a class. As such, each sample can be reconstructed more accurately by the samples of the same class as much as possible. Thus, we introduce a block-diagonal structure constraint matrix $Q$ to constrain the coefficient matrix by minimizing the approximation error between $Q$ and $W_0 ... W_M(Z_0 ... Z_M)^T A^T$ in each layer:

$$J_2 = \left\|Q - W_0 ... W_M \left(Z_0 ... Z_M\right)^T A^T\right\|_F^2 + \left\|W_0 ... W_M \left(Z_0 ... Z_M\right)^T A^T\right\|_F^2, \tag{14}$$

where the structure constraint matrix $Q$ is defined as follows:

$$Q = \begin{bmatrix} Q_L & 0 \\ 0 & Q_U \end{bmatrix} \in \mathbb{R}^{(l+u)\times(l+u)},\ Q_L = \begin{bmatrix} Q_1 & 0 & 0 & 0 \\ 0 & Q_2 & 0 & 0 \\ 0 & 0 & ... & 0 \\ 0 & 0 & 0 & Q_c \end{bmatrix} \in \mathbb{R}^{l\times l}, \tag{15}$$

where $Q_L$ and $Q_U$ are the structure constraint matrices defined based on the labeled data $X_L$ and unlabeled data $X_U$. Since the samples of $X_L$ are originally labeled, $Q_L$ is a strict block-diagonal matrix, where each block $Q_i$ ($i=1,2,...,c$) is an $l_i \times l_i$ matrix of all ones, defined according to the labeled samples, and $l_i$ is the number of samples in class $i$ in $X_L$. For example, if we have 9 labeled samples, where $x_1$ and $x_2$ are from the class 1, $x_3$, $x_4$, $x_5$ and $x_6$ are from class 2 and the remaining ones are from class 3, the sub-matrices $Q_1$, $Q_2$ and $Q_3$ can be defined as

$$Q_1 = \begin{bmatrix} 1 & 1 \\ 1 & 1 \end{bmatrix},\ Q_2 = \begin{bmatrix} 1 & 1 & 1 & 1 \\ 1 & 1 & 1 & 1 \\ 1 & 1 & 1 & 1 \\ 1 & 1 & 1 & 1 \end{bmatrix},\ Q_3 = \begin{bmatrix} 1 & 1 & 1 \\ 1 & 1 & 1 \\ 1 & 1 & 1 \end{bmatrix}.$$

It should be noted that we initiate $Q_U$ by the cosine similarities over the samples in $X_U$ and update $Q_U$ in $M$-th ($M>1$) layer using the cosine similarity matrix defined on the new representation of $X_U$, i.e., $(V_m)_i$, $i \in \{l+1,...,N\}$. In this way, we can ensure the overall coefficient matrix $W_0 ... W_M(Z_0 ... Z_M)^T A^T$ to have a good structure for the representation learning.

## C. Self-weighted Dual-graph Learning

To obtain the locality preserving representation, we further add a self-weighted dual-graph learning into DS²CF-Net, which can preserve the neighborhood information of both the deep basis vectors $XW_0...W_M$ and representations $(Z_0...Z_M)^T A^T$ in an adaptive manner at the same time. Specifically, we compute the data weight matrix $S^V \in \mathbb{R}^{N\times N}$ over the deep representations and the feature weight matrix $S^U \in \mathbb{R}^{D\times D}$ over the deep basis vectors adaptively by solving the following reconstructive loss:

$$J_3 = \left\| \left(XW_0...W_M\right)^T - \left(XW_0...W_M\right)^T S^U \right\|_F^2 + \left\| \left(\left(Z_0...Z_M\right)^T A^T\right) - \left(\left(Z_0...Z_M\right)^T A^T\right) S^V \right\|_F^2, \; s.t.\; S^U \geq 0, S^V \geq 0 \quad (16)$$

Clearly, the nonnegative dual-graph weights of DS²CF-Net are different from those of GCF [13] in two aspects. First, GCF is a single-layer model that defines the weights over the "shallow" basis vectors and features, while our DS²CF-Net encodes the locality based on the deep basis vectors and features. Second, our DS²CF-Net does not need to specify the number of nearest neighbors, suffered in GCF, since the neighbors of each sample are determined automatically in DS²CF-Net by directly minimizing the reconstruction error. In addition, the dual-graph weights are adaptive, and are also updated with the factorization process, which can enable DS²CF-Net to be adaptive to different datasets and produce accurate feature representations.

**Fine-tuning of the structure-constrained matrix $Q_U$ and data weight matrix $S^V$ using $A_U$.** In this process, we consider how to refine $Q_U$ and $S^V$ after obtaining the soft labels $A_U$ for $X_U$, so that the learned representations are better. Specifically, in each iteration, we first obtain the hard labels $A_U^*$ by setting the maximum value in each column of $A_U$ to 1 and otherwise setting it to 0. The fine-tuning process can then be performed as follows: if $\left(A_U^*\right)_{ij} = 0$, then $(Q_U)_{ij}=0$ and $(S^V)_{ij}=0$. This is easy to understand, since $\left(A_U^*\right)_{ij} = 0$ means that the unlabeled samples $x_i$ and $x_j$ are not in the same class. Hence, the structure constraint information and the data weight between them shall be ideally zeros for the consideration of discrimination.

## D. Error Correction Mechanism and Feature Fusion

For a multi-layer model, it is important to ensure that the performance will not decrease fast with the increasing number of layers. In order to avoid this potential risk for our DS²CF-Net, we clearly incorporate an error correction mechanism and a feature fusion strategy into our framework. Specifically, between two consecutive layers, we include a clustering evaluation module to evaluate the clustering performance over the features in each layer. Specifically, for the new representation $V_i$ in the $i$-th ($i>1$) layer, we input it into the clustering evaluation module, i.e., we perform the clustering evaluations by K-means algorithm based on $V_i$, and then we can obtain the clustering accuracy $AC_i$. To avoid fast degradation as the number of layers increases, the error correction and feature fusion are performed as follows. If $AC_i$ is larger than $AC_{i-1}$, then directly moving to the $(i+1)$-th layer, otherwise performing the feature fusion over $V_i$, i.e., adding $V_i$ and the features of the first $i-1$ layers together to update $V_i = V_1+\cdots+V_i$. In this way, DS²CF-Net can deliver more stable and reliable results in the multi-layer case. Note that these strategies will be verified by extensive simulations.

## E. Objective Function

Based on the above analysis, the final objective function of our DS²CF-Net method can be formulated as

$$\begin{aligned} O_{DS^2CF-Net} = \min_{\substack{W_1,...,W_M, Z_1,...,Z_M,\\ S^U, S^V, P}} & \left\| X - XW_0...W_M \left(Z_1...Z_M\right)^T A^T \right\|_F^2 \\ & + \alpha\left[\left\|Q - R_M\right\|_F^2 + \left\|R_M\right\|_F^2\right] + \beta\left[\left\|U_M^T - U_M^T S^U\right\|_F^2 + \left\|V_M^T - V_M^T S^V\right\|_F^2\right] \\ & + \gamma\left[\left\|A_L - X_L^T P\right\|_F^2 + \left\|P^T X - P^T X R_M\right\|_F^2 + \left\|P\right\|_{2,1}\right] \\ s.t.\; & \forall_{m\in\{1,2,...,M\}} W_m \geq 0, Z_m \geq 0, S^U \geq 0, S^V \geq 0 \end{aligned} ,(17)$$

where $U_M = XW_0 \cdots W_M$, $V_M = A(Z_0...Z_M)$ and $R_M = W_0...W_M V_M^T$.

# IV. Optimization and Computational Complexity

## A. Optimization

From the objective function of DS²CF-Net, we can easily find that the involved variables $W_m$, $Z_m$ ($m\in\{1,2,...,M\}$), $S^U$, $S^V$ and $P$ depend on each other, so they cannot be solved directly. Following the common procedures, we present an iterative optimization strategy using the Multiplicative Update Rules (MUR) method [44-45] for obtaining local optimal solutions. Specifically, we solve the problem by updating the variables alternately and optimize one of them each time by fixing the others. The detailed optimization procedures are showed as follows:

### 1) Fix others, update the factors $W_m$ and $Z_m$:

We first show how to optimize $W_m$ and $Z_m$. For the $m$-th layer, $W_1,\ldots,W_{m-1}$, $Z_1,\ldots,Z_{m-1}$ and $P$ are all known as the constants. By defining $\Pi_{m-1} = W_0...W_{m-1}$ and $\Lambda_{m-1} = Z_0...Z_{m-1}$, the reduced subproblem associated with $W_m$ and $Z_m$ can be defined as

$$\begin{aligned} \min_{W_m, Z_m} & \left\| X - X\Pi_{m-1}W_m\left(\Lambda_{m-1}Z_m\right)^T A^T \right\|_F \\ & + \alpha\left(\left\|Q - R_M\right\|_F^2 + \left\|R_M\right\|_F^2\right) + \beta\left(\left\|U_M^T - U_M^T S^U\right\|_F^2 + \left\|V_M - V_M S^V\right\|_F^2\right) \\ & + \gamma\left\|P^T X - P^T X R_M\right\|, \; s.t.\; W_m, Z_m \geq 0 \end{aligned} , (18)$$

where $U_M = X\Pi_{m-1}W_m$, $V_M = A\Lambda_{m-1}Z_m$ and $R_M = \Pi_{m-1}W_m\left(\Lambda_{m-1}Z_m\right)^T A^T$. Let $\psi_{ik}^w$ and $\psi_{ik}^z$ be the Lagrange multipliers for the constraints $(W_m)_{ik} \geq 0$ and $(Z_m)_{ik} \geq 0$, $\Psi_w = \left[\psi_{ik}^w\right]$ and $\Psi_z = \left[\psi_{ik}^z\right]$, then the Lagrange function can be constructed as

$$\begin{aligned} \wp = & \left\| X - X\Pi_{m-1}W_m\left(\Lambda_{m-1}Z_m\right)^T A^T \right\|_F \\ & + \alpha\left[\left\|Q - \Pi_{m-1}W_m\left(\Lambda_{m-1}Z_m\right)^T A^T\right\|_F^2 + \left\|\Pi_{m-1}W_m\left(\Lambda_{m-1}Z_m\right)^T A^T\right\|_F^2\right] \\ & + \beta tr\left[\left(X\Pi_{m-1}W_m\right)^T H_u \left(X\Pi_{m-1}W_m\right) + \left(A\Lambda_{m-1}Z_m\right)^T H_v \left(A\Lambda_{m-1}Z_m\right)\right] \\ & + \gamma\left\|P^T X - P^T X\Pi_{m-1}W_m\left(\Lambda_{m-1}Z_m\right)^T A^T\right\| + tr\left(\Psi_w W_m^T\right) + tr\left(\Psi_z Z_m^T\right) \end{aligned} ,(19)$$

where $H_u = \left(I - S^U\right)\left(I - S^U\right)^T$, $H_v = \left(I - S^V\right)\left(I - S^V\right)^T$ and $I$ denotes an identity matrix. Then, $W_m$ and $Z_m$ can be alternately updated by fixing others. Let $K_X = X^T X$, $K_A = A^T A$ and $K_P = X^T P P^T X$, the derivatives w.r.t. $W_m$ and $Z_m$ can be obtained as

$$\begin{aligned}\partial \wp/\partial W_m &= 2\left(\Pi_{m-1}{}^T K_X \Pi_{m-1} W_m V_M^T V_M - \Pi_{m-1}{}^T K_X V_M\right) \\ &+ 2\alpha\left(-\Pi_{m-1}{}^T Q V_M + 2Q^T \Pi_{m-1} W_m V_M^T V_M\right) \\ &+ \beta\left(\Pi_{m-1}{}^T X^T \left(H_u + H_u{}^T\right) X \Pi_{m-1} W_m\right) \\ &+ 2\gamma\left(\Pi_{m-1}{}^T K_P \Pi_{m-1} W_m V_M^T V_M - \Pi_{m-1}{}^T K_P V_M\right) + \Psi_w\end{aligned} \quad (20)$$

$$\begin{aligned}\partial \wp/\partial Z_m &= 2\left(\Lambda_{m-1}^T K_A \Lambda_{m-1} Z_m U_M^T U_M - \Lambda_{m-1}^T A^T K_X \Pi_m\right) \\ &+ 2\alpha\left(2\Lambda_{m-1}^T K_A \Lambda_{m-1} Z_m W_m{}^T Q^T \Pi_m - \Lambda_{m-1}^T A^T Q^T \Pi_m\right) \\ &+ \beta\left(\Lambda_{m-1}^T \left(H_v + H_v{}^T\right) \Lambda_{m-1} Z_m K_A^T\right) \\ &+ 2\gamma\left(\Lambda_{m-1}^T K_A \Lambda_{m-1} Z_m U_M^T P P^T U_M^T - \Lambda_{m-1}^T A^T K_P \Pi_m\right) + \Psi_z\end{aligned} , \quad (21)$$

where $\Pi_m = \Pi_{m-1} W_m$, and $\Pi_m$ is known when updating $Z_m$. By using the KKT conditions $\psi_{ik}^w (W_m)_{ik} = 0$ and $\psi_{ik}^z (Z_m)_{ik} = 0$, we can obtain the updating rules for $W_m$ and $Z_m$:

$$\left(W_m\right)_{ik} \leftarrow \left(W_m\right)_{ik} \frac{\left(2\Pi_{m-1}^T K_X V_M + 2\alpha \Pi_{m-1}^T Q V_M + \Omega_W\right)_{ik}}{\left(2\Pi_{m-1}^T K_X \Pi_{m-1} W_m V_M^T V_M + \Phi_W\right)_{ik}}, \quad (22)$$

$$\left(Z_m\right)_{ik} \leftarrow \left(Z_m\right)_{ik} \frac{\left(2\Lambda_{m-1}^T A^T K_X \Pi_m + 2\alpha \Lambda_{m-1}^T A^T Q^T \Pi_m + \Omega_Z\right)_{ik}}{\left(2\Lambda_{m-1}^T K_A \Lambda_{m-1} Z_m U_M{}^T U_M + \Phi_Z\right)_{ik}}, \quad (23)$$

where $\Phi_W = 4\alpha Q^T \Pi_{m-1} W_m V_M^T V_M + \beta \Pi_{m-1}{}^T X^T \left(H_u + H_u{}^T\right) X \Pi_{m-1} W + 2\gamma \Pi_{m-1}{}^T K_P \Pi_{m-1} W_m V_M^T V_M$, $\Phi_Z = 4\alpha \Lambda_{m-1}^T K_A \Lambda_{m-1} Z_m W_m{}^T Q^T \Pi_m + \beta \Lambda_{m-1}^T \left(H_v + H_v{}^T\right) \Lambda_{m-1} Z_m K_A^T + 2\gamma \Lambda_{m-1}^T K_A \Lambda_{m-1} Z_m U_M^T P P^T U_M$, $\Omega_W = 2\gamma \Pi_{m-1}^T K_P V_M$ and $\Omega_Z = 2\gamma \Lambda_{m-1}^T A^T K_P \Pi_m$ are auxiliary matrices.

### 2) Fix others, update the weight matrices $S^U$ and $S^V$:

When other variables are computed, we can use them to update the dual-graph weights $S^U$ and $S^V$ by removing the irrelevant terms to $S^U$ and $S^V$ from the objective function. Let $\vartheta_{ik}^u$ and $\vartheta_{ik}^v$ denote the Lagrange multipliers for the constraints $S_{ik}^U \geq 0$ and $S_{ik}^V \geq 0$, $\Omega_u = \left[\vartheta_{ik}^u\right]$ and $\Omega_v = \left[\vartheta_{ik}^v\right]$, then the Lagrange function of the reduced problem can be similarly defined as

$$\wp = \beta\left(\left\|U_M^T - U_M^T S^U\right\|_F^2 + \left\|V_M^T - V_M^T S^V\right\|_F^2\right) + tr\left(\Omega_u S^{U^T}\right) + tr\left(\Omega_v S^{VT}\right), \quad (24)$$

where $U_M = X\Pi_{m-1} W_m$ and $V_M = A\Lambda_{m-1} Z_m$ are known variables in this step. Based on the KKT conditions $\vartheta_{ik}^u S_{ik}^U = 0$ and $\vartheta_{ik}^v S_{ik}^V = 0$, we can obtain the updating rules for $S^U$ and $S^V$:

$$\left(S^U\right)_{ik} \leftarrow \left(S^U\right)_{ik} \frac{\left(\left(X\Pi_{m-1} W_m\right)\left(X\Pi_{m-1} W_m\right)^T\right)_{ik}}{\left(\left(X\Pi_{m-1} W_m\right)\left(X\Pi_{m-1} W_m\right)^T S^U\right)_{ik}}, \quad (25)$$

$$\left(S^V\right)_{ik} \leftarrow \left(S^V\right)_{ik} \frac{\left(\left(A\Lambda_{m-1} Z_m\right)\left(A\Lambda_{m-1} Z_m\right)^T\right)_{ik}}{\left(\left(A\Lambda_{m-1} Z_m\right)\left(A\Lambda_{m-1} Z_m\right)^T S^V\right)_{ik}}. \quad (26)$$

### 3) Fix others, update the robust label predictor $P$:

Finally, we solve the projection $P$ from Eq.(17), with $W_m$, $Z_m$, $S^U$ and $S^V$ known. By the properties of $L_{2,1}$-norm [38-41], we have $\|P\|_{2,1} = 2tr\left(P^T B P\right)$, where $B$ is a $D \times D$ diagonal matrix with entries $b_{ii} = 1/\left(2\left\|p^i\right\|_2\right)$, where $p^i$ is the $i$-th row of $P$. Thus, we can infer the label predictor $P$ from the following problem:

**Algorithm 1: Optimization procedures of DS²CF-Net**

**Inputs:** Partially labeled data matrix $X=[X_L, X_U]$, constant $r$ and tunable parameters $\alpha, \beta, \gamma$.

**Initializations:**

Initialize $W$ and $Z$ to be the random matrices;

Initialize the diagonal matrix $B$ as the identity matrix;

Initialize the linear label predictor as $P = \left(X_L X_L{}^T + I\right)^{-1} X_L A_L$, use $P$ to predict the soft labels of unlabeled data as $X_U{}^T P$, and then apply the normalized soft labels by Eq.(13) to initialize the label constraint matrix $A$ by using Eq.(9);

Initialize $Q_U$ by the cosine similarities over $X_U$;

Initialize $S^U$ using the cosine similarities over $X$ and initialize $S^V$ using the semi-supervised weights, i.e., supervised ones for $X_L$ and the cosine similarities for $X_U$;

Initial clustering accuracy $AC_0=0$; $t=0$; Initialize $m$=1.

***For each fixed number m of layers:***

*While not converged do*

1. Update the matrix factors $W_m^{t+1}$ and $Z_m^{t+1}$ by Eqs.(22-23), and then we can obtain $V_m^{t+1} = AZ_0 ... Z_m^{t+1}$;
2. Update the weights $\left(S^U\right)^{t+1}$ and $\left(S^V\right)^{t+1}$ by Eqs.(25-26);
3. Update the linear label predictor $P^{t+1}$ by Eq.(28), update the soft labels of $X_U$ as $X_U{}^T P^{t+1}$, and then update $A_U$ by Eq.(13);
4. Update the full label-constraint matrix $A$ by Eq.(9);
5. Update $Q_U$ by the cosine similarities defined based on $\left(V_m^{t+1}\right)_i$, $i \in \{l+1, ..., N\}$, and update the structure-constraint matrix $Q$;
6. Fine-tuning of $S^V$ and $Q$ based on the soft labels $A_U$;
7. *Check for convergence*: if $\left\|W_m^{t+1} - W_m^t\right\|_F^2 \leq \varepsilon$ and $\left\|V_m^{t+1} - V_m^t\right\|_F^2 \leq \varepsilon$, stop; else *t=t+1*.

*End while*

8. Clustering evaluation on $V_m$ by K-means and obtain $AC_m$;
9. *Error correction and feature fusion:* If $AC_m < AC_{m-1}$, then obtaining the fused features as $V_m = V_m + ... + V_1$ ; else go to step 1.

***End for***

**Output:** Deep representation $V_m^*$ and clustering result $AC_m$.

$$\min_{P, B} \left\|A_L - X_L^T P\right\|_F^2 + \left\|P^T X \left(I - R_M\right)\right\|_F^2 + tr\left(P^T B P\right), \quad (27)$$

where each $p_i \neq 0$. By seeking the derivative of the above problem *w.r.t.* $P$, we can infer $P$ in each layer as follows:

$$P = \left(X_L X_L^T + X H_M X^T + B\right)^{-1} X_L A_L, \quad (28)$$

where $H_M=(I-R_M)(I-R_M)^T$. After $P$ is obtained, we can use it to update the diagonal matrix $B$ and predict the labels of unlabeled samples. After that, we can use the normalized soft labels to optimize the label constraint matrix $A$ for representation.

For complete presentation, we summarize the optimization procedures of DS²CF-Net in Algorithm 1, where the diagonal matrix $B$ is initialized as an identity matrix. We initialize the linear label predictor $P = \left(X_L X_L{}^T + I\right)^{-1} X_L A_L$ as [42] and predict the soft labels of unlabeled data as $X_U{}^T P$, and then we normalize the soft labels by Eq.(13). Based on the normalized soft labels of unlabeled data, we can initialize the label constraint matrix $A$. Since DS²CF-Net jointly optimizes the basis vectors and representation matrices that are two major variables, to ensure the proposed algorithm to converge, the stopping condition can be simply set to $\left\|W_m^{t+1} - W_m^t\right\|_F^2 \leq \varepsilon$ and $\left\|V_m^{t+1} - V_m^t\right\|_F^2 \leq \varepsilon$ ($\varepsilon = 10^{-3}$) in the $m$-th layer, where $V_m^{t+1} = AZ_0 ... Z_m^{t+1}$ is the computed representation matrix in the $m$-th layer and the approximation errors measure the difference between two sequential sets of basis vectors and representation matrices, which can make sure that the representation learning result will not change drastically. Note that an early version was presented in [54]. This paper has also incorporated an error analysis mechanism and a feature fusion strategy, so that more stable and reliable representation and clustering results can be obtained. In addition, a new fine-tuning process is performed to refine the structure-constrained matrix and data weight matrix in each layer for obtaining more accurate representations. Besides, this paper also provided the time complexity analysis and conducted a thorough evaluation on the tasks of representation learning and clustering by including visual image analysis and adding more real-world databases.

### B. Computational Complexity Analysis

We discuss the time complexity of DS$^2$CF-Net. We use the big **O** notation to show the complexity of our algorithm as [53]. For each layer, according to the updating rules of our DS$^2$CF-Net, we need to perform the extra computation of $A_U$, $S^U$, $S^V$, $Q$, and $P$ over the CCF algorithm. We can find that the big **O** of each updating operation for each variable is not more than $\mathbf{O}(N^3)$ in the optimization procedure if $N$ is larger than the dimensionality $D$. Since the number of layers and the number of iteration times are all constants, they are negligible when calculating the computational complexity. Overall, the time complexity of our DS$^2$CF-Net is $\mathbf{O}(N^3)$. Note that we also report the actual runtime performance comparison to other methods in Section V.

## V. Experimental Results and Analysis

In this section, we mainly conduct simulations to examine the data representation and clustering performance of our DS$^2$CF-Net. The experimental results of our DS$^2$CF-Net are compared with those of 5 deep ML models (i.e., MNMF [18], MCF [19], GMCF [21], DSNMF [23] and DSCF-Net [43]), 3 single-layer MF models (i.e., DNMF [12], GCF [13] and SRMCF [24]), and four semi-supervised MF models (i.e., SemiGNMF [6], CNMF [16], CCF [17] and RS$^2$ACF [42]). Note that SemiGNMF adds class information of labeled data into the graph structures by modifying the graph weight matrix [6][17]. In this study, 6 public image databases are involved, including two face image databases (i.e., AR [25] and MIT CBCL [26]), two object image databases (i.e., COIL100 [27] and ETH80 [28]), three handwritten image datasets (i.e., USPS [29], EMNIST Letters [53] and EMNIST Digits [53]), and one fashion products database (i.e., Fashion MNIST [46]). Some sample images of the evaluated datasets are shown in Fig.3, and the detailed information about the used databases are described in Table 1, where we show the total number of samples, dimension and number of classes. For each face or object image, we follow the common procedures [30-31] to resize it into 32×32 pixels, forming a 1024-dimensional sample vector. Finally, we can obtain a data matrix with the vectorized representations of all the images as columns. The vectorized process for USPS, EMNIST Letters, EMNIST Digits and Fashion MNIST databases are similar. In this work, we normalize each column of the input data matrix to have unit norm for each database. We perform all experiments on a PC with Intel Core i5-4590 CPU @ 3.30 GHz 3.30 GHz 8G.

**Table 1:** List of evaluated databases and database information.

| Data Type | Name | #sample | #class | #dim |
|---|---|---|---|---|
| **Face images** | AR [25] | 2600 | 100 | 1024 |
| | MIT CBCL [26] | 3240 | 10 | 1024 |
| **Object images** | COIL100 [27] | 7200 | 100 | 1024 |
| | ETH80 [28] | 3280 | 80 | 1024 |
| **Handwriting** | USPS [29] | 9298 | 10 | 256 |
| | EMNIST Letters [53] | 145600 | 26 | 784 |
| | EMNIST Digits [53] | 280000 | 10 | 784 |
| **Fashion products** | Fashion MNIST [46] | 70000 | 10 | 784 |

### A. Visual Image Analysis by Visualization

***Visualization of the adaptive weight matrix $S^V$.*** The obtained representation $V_M = A(Z_0...Z_M)$ is the final output of our model, we first evaluate the representation ability of $V_M$ by visualizing the adaptive weights $S^V$ on $V_M$ . AR face database and COIL100 object database are used in this study. For the AR database, we randomly choose 2 categories to construct the adjacency graph $S^V$ for clear observation, with 10 labeled images per class (that is, 20 labeled samples and 32 unlabeled samples in total). For COIL100 database, we randomly choose 4 categories to construct $S^V$, with 28 labeled samples per category (i.e., 28 labeled and 44 unlabeled). The weight matrices $S^V$ are shown in Fig.4, where we show the adaptive weights obtained by DS$^2$CF-Net in the first four layers on each database. Note that the green box contains the weights on labeled data and the yellow box contains those on unlabeled data. We see that the weight matrices have approximate block-diagonal structures in each layer. Specifically, the structures of the adaptive weights get clearer with less noise and inter-class connections as the number of layers increases, which means that the learned new representation $V_M$ has a strong representation ability and moreover our deep model can potentially improve the similarity measure.

***Visualization of the structure constraint matrix Q.*** Since the structure constraint matrix $Q$ determines the structures of the self-expressive coefficient matrix to encode the smoothness of manifolds in the process of label propagation, we also visualize its structures for observation. AR face database and COIL100 database are used. For AR, we randomly choose 2 categories for the test with 10 labeled images per class; For COIL100, we randomly choose 4 categories for the test, with 28 labeled samples per category. The structure constraint matrices $Q$ are shown in Fig.5, where we show the results obtained in the first four layers over each database, the green box and the yellow box contains the parts $Q_L$ and $Q_U$ on the labeled and unlabeled data, respectively. $Q_L$ is defined according to the known labels, while $Q_U$ is the cosine similarity defined based on the new representation of unlabeled samples. We find that the constraint matrix $Q$ has a clear block-diagonal structure, and moreover the structures become better with the increasing number of layers, which implies that the structure constraint matrix $Q$ in each layer has a strong discriminative representation power.

### B. Convergence Analysis and Runtime Comparison

***(1) Convergence Analysis.*** The involved variables of our algorithm are optimized alternately in each layer, while the actual runtime performance of each iterative algorithm is closely related to the number of iterations in reality, so we would like to present some convergence analysis results. The MIT CBCL and Fashion MNIST databases, with 40% samples labeled, are used

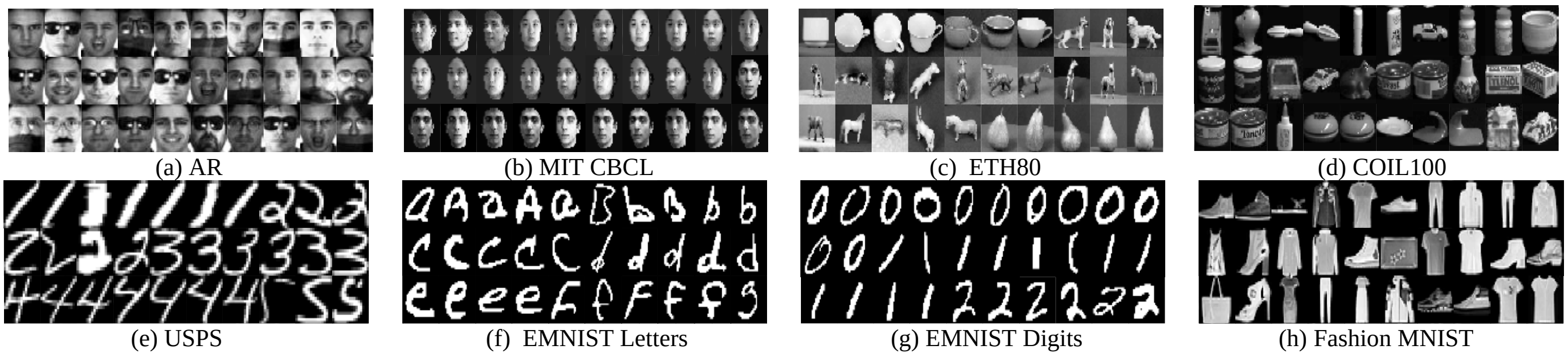

**Figure 3:** Sample images of the evaluated real image databases.

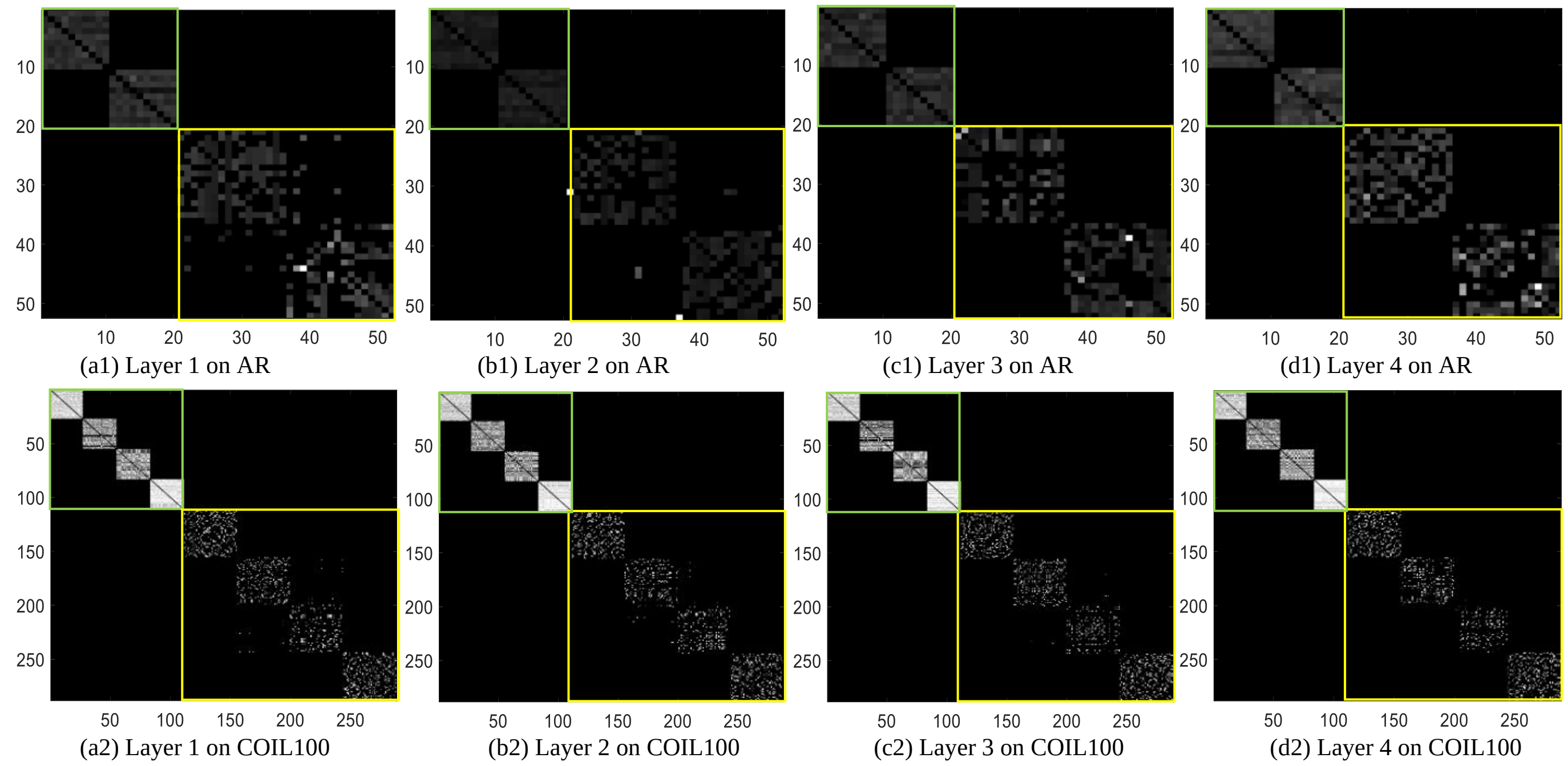

**Figure 4:** Visualization of the data weight matrix $S^V$ obtained by our DS$^2$CF-Net in the first four layers over AR and COIL100 databases.

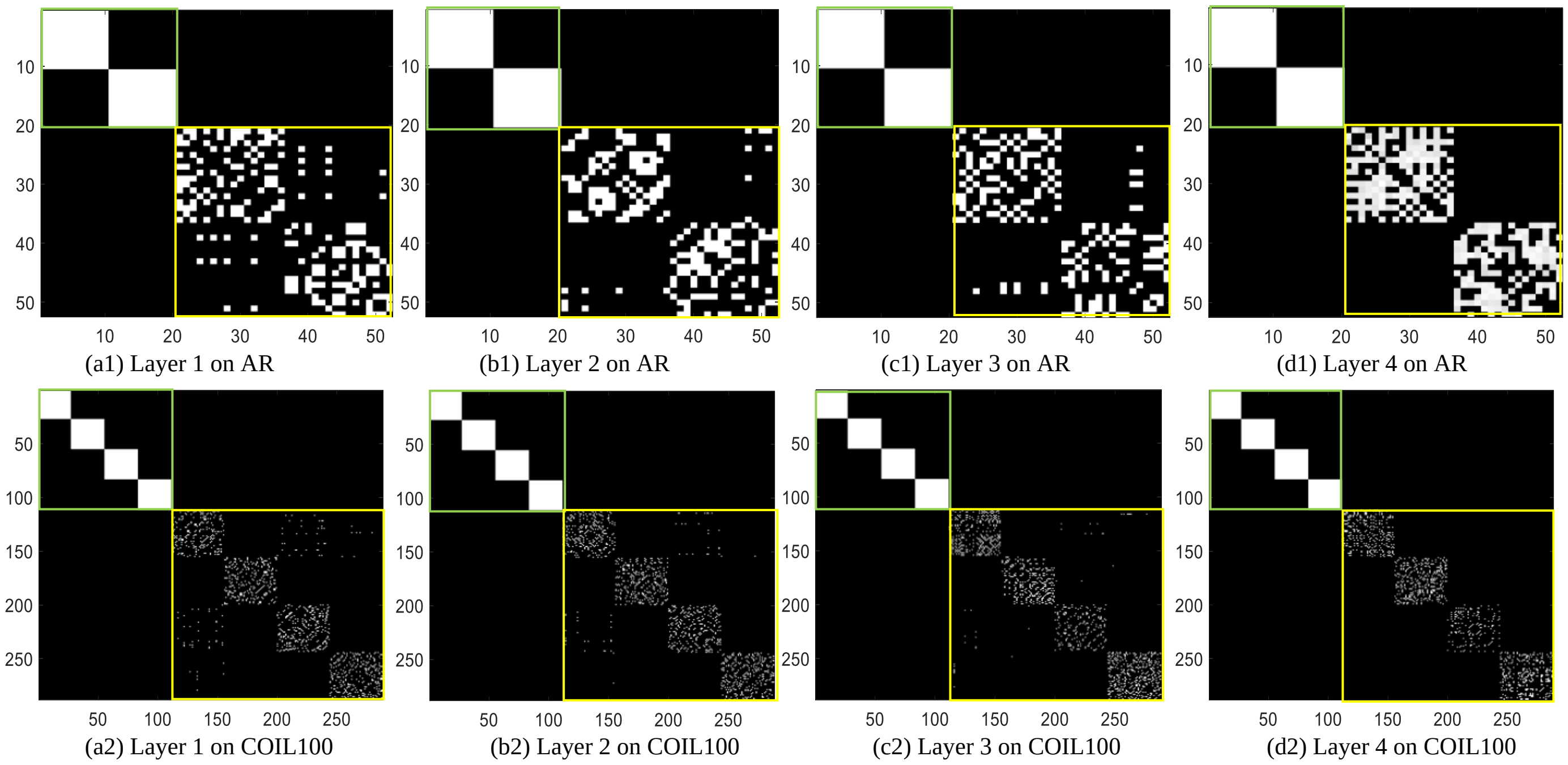

**Figure 5:** Visualization of the structure constraint matrix $Q$ obtained by our DS$^2$CF-Net in the first four layers over AR and COIL100 databases.

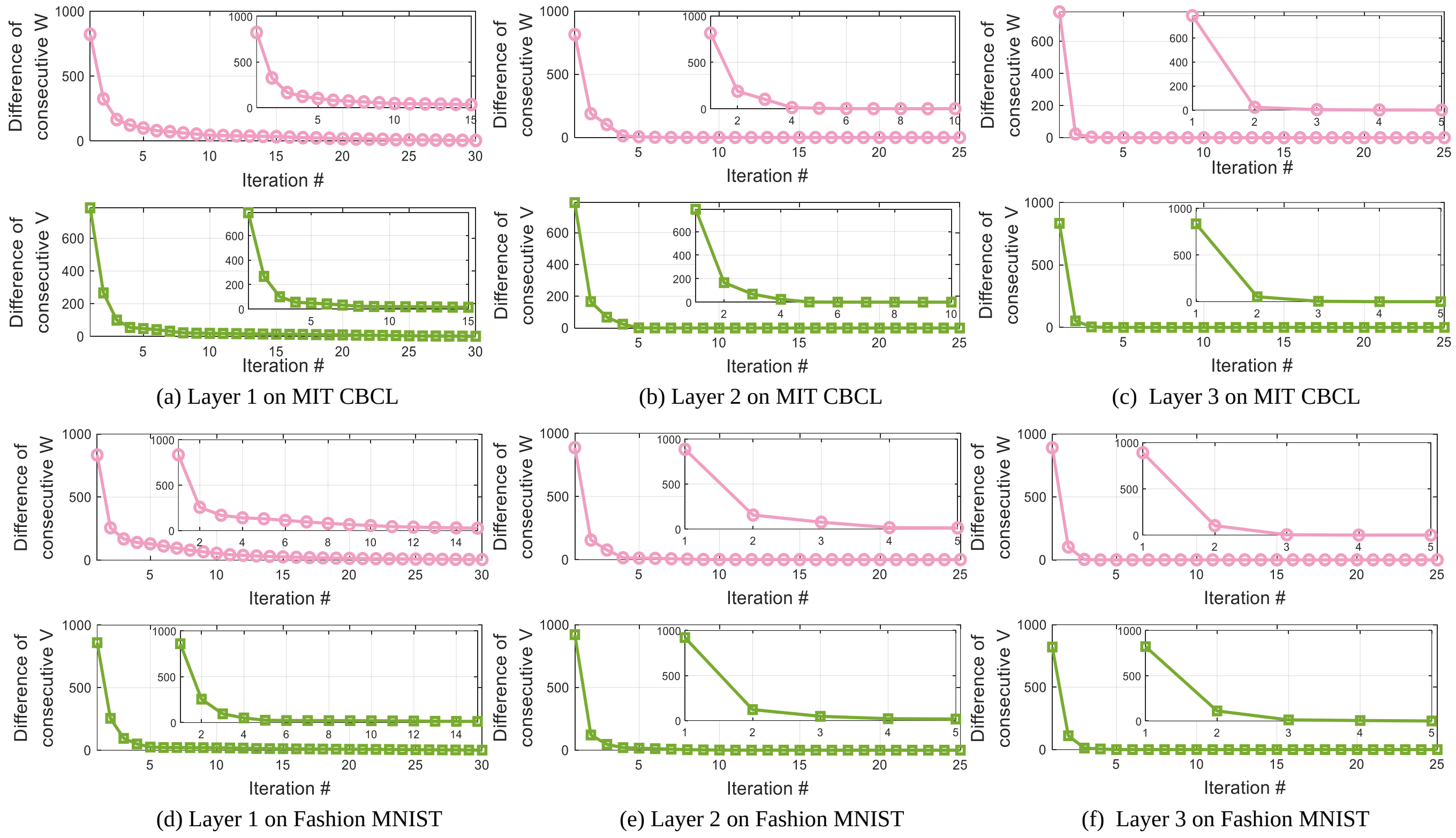

**Figure 6:** Convergence analysis of our proposed DS$^2$CF-Net algorithm on the MIT CBCL face database and Fashion MNIST database.

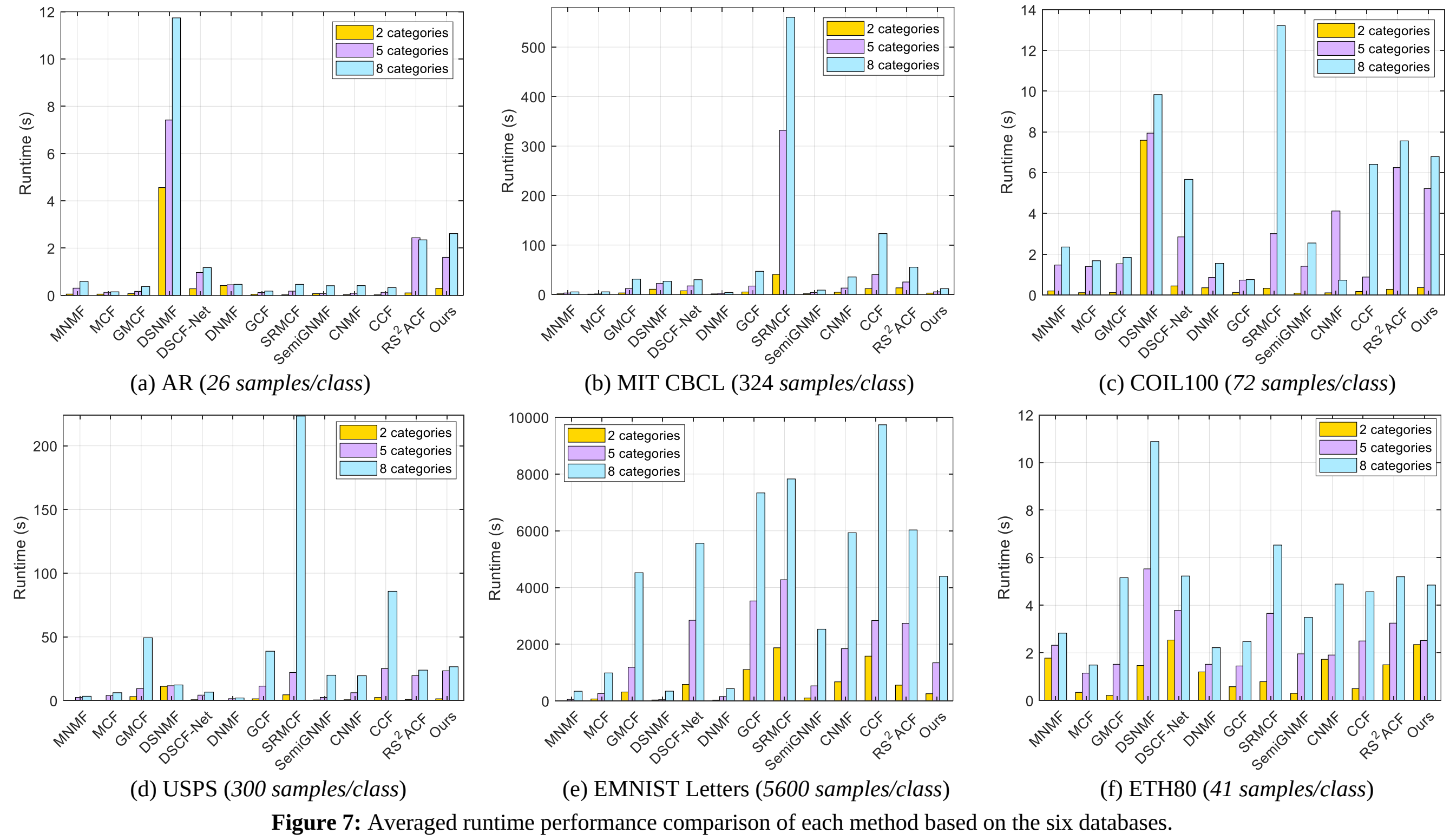

**Figure 7:** Averaged runtime performance comparison of each method based on the six databases.

to train our method in this study. We show the convergence results of our DS$^2$CF-Net in the first three layers in Fig.6, where the X-axis shows the number of iterations and Y-axis denotes the difference between two consecutive basis vectors (i.e., $W^{t+1}$ and $W^{t}$) and two consecutive representations (i.e., $V^{t+1}$ and $V^{t}$) respectively, i.e., $\|W^{t+1}-W^{t}\|_F^2$ and $\|V^{t+1}-V^{t}\|_F^2$. We see that: 1) DS$^2$CF-Net converges rapidly in each layer; 2) with the increasing number of layers, our DS$^2$CF-Net converges more rapidly

due to the effects of deep representation, which usually converges within 5 iterations in the 3rd layer. Note that as a multi-layer model, this is beneficial for the efficiency.

***(2) Runtime Comparison.*** In addition to presenting the computational time complexity analysis of our DS$^2$CF-Net, we also would like to show the actual running time of each method (in second and averaged based on 10 runs) for the fair comparison. It is worth noting that we have compared 6 multi-layer methods, so to facilitate the comparison, we report the averaged runtime of different layers for these multi-layer models. In this study, we employ six databases, including AR, MIT CBCL, COIL100, USPS, EMNIST Letters, and ETH80 for evaluations. For each database, we randomly select 2, 5 and 8 categories to train each model. The runtime performance comparison results are given in Fig.7. We see that: 1) the needed running time is increased with the increasing number of samples, i.e., from small-scale to large-scale; 2) DSNMF needs more time for AR, ETH80 and COIL100 databases, while needing less time for large-scale databases relatively, since it spends most of the running time in initialization. SRMCF needs the most time for several databases. CNMF, CCF and GCF also needs more time than other methods in most cases, especially on the large-scale databases. The main reason may be because they need more time for the convergence of the algorithm; 3) our proposed DS$^2$CF-Net needs comparable time to RS$^2$ACF based on each database, which spend slightly more time than other methods. Overall, in most cases the actual running time of our DS$^2$CF-Net is acceptable due to fast convergence, although it is a multi-layer model.

## C. Quantitative Clustering Evaluation

***(1) Clustering evaluation process.*** For the quantitative clustering evaluations, we perform the K-means algorithm with cosine distance on the representation obtained by each model. Following the procedures in [17] [34], for each number K of clusters, we choose K categories from each database randomly and use the samples of K categories to form the data matrix $X$. The value of K is tuned from 2 to 10 in our study. The rank of the representation is set to K+1 for clustering as [17]. The clustering results are averaged based on 10 random selections of the K categories. For fair comparison, we choose 40% labeled samples per class for each semi-supervised algorithm (i.e., SemiGNMF, CNMF, CCF, RS$^2$ACF and our DS$^2$CF-Net). For fair comparison to the existing multi-layer matrix factorization methods (i.e., MNMF, MCF, GMCF, DSNMF and DSCF-Net), we report the highest clustering scores in their first 10 layers of each method, rather than fixing the number of layers.

***(2) Clustering evaluation metric.*** We employ two widely-used evaluation methods, i.e., *Accuracy* (AC) and *F-measure* [35-36]. AC is the percentage of the cluster labels to the true labels provided by the original data corpus, defined as follows:

$$AC = \left[\sum_{i=1}^{N} \delta\left(r_i, map\left(p_i\right)\right)\right] / N \,, \tag{29}$$

where $N$ is the number of samples, and the function $map\left(p_i\right)$ is the permutation mapping function that maps the cluster label $p_i$ obtained by the clustering method to the true label $r_i$ provided by the data corpus, and the best mapping can be obtained by the Kuhn-Munkres algorithm [37] according to [35]. The clustering F-measure is defined as follows:

$$F_\mu = \frac{\left(\mu^2+1\right) PRECISION \times RECALL}{\mu^2 PRECISION + RECALL} \,, \tag{30}$$

where we set the parameter $\mu = 1$. Note that both values of the AC and F-measure range from 0 to 1, i.e., the higher the value is, the better the clustering result will be.

***(3) Clustering evaluation results***

**Face Clustering.** We first use face images to evaluate the clustering ability of learned representation by each method. AR and MIT CBCL face database are evaluated. The clustering performance in terms of AC and F-measure over varied K numbers is tested. The clustering curves on the AR and MIT CBCL databases are shown in Figs.8(a-b), respectively. The averaged AC and F-scores according to the curves in Figs.8(a-b) are summarized in Tables 2-3, respectively. We can see that: (1) the obtained AC and F-measure of each method go down as the number of categories is increased, which is easy to understand, since clustering data of less categories is relatively easier than clustering more categories; (2) our DS$^2$CF-Net delivers higher values of AC and F-measure than other compared methods in the investigated cases. Both RS$^2$ACF and DSCF-Net performs better than other remaining methods in most cases. CNMF also delivers promising results over the MIT CBCL database.

**Object Clustering.** We then evaluate each method for representing and clustering the object image data. In this experiment, COIL100 and ETH80 object databases are evaluated. The clustering curves on ETH80 and COIL100 databases are shown in Figs.8(c-d), respectively. The averaged AC and F-scores according to the curves in Figs. 8(c-d) are summarized in Tables 2-3, respectively. We can see that the increasing number of selected categories clearly decreases the performance of each method due to the fact that clustering data of less categories is relatively easier. It can also be found that DS$^2$CF-Net delivers higher values of AC and F-measure than other evaluated methods in most cases. In addition, the semi-supervised SemiGNMF, CNMF, CCF and RS$^2$ACF and methods can perform better than other algorithms using partially labelled data, where RS$^2$ACF is the best method in this case. DSCF-Net also delivers relatively better clustering performance than other multilayer methods.

**Handwritten Digit Clustering.** We also examine the performance of each method for clustering the handwritten digits of the USPS, EMNIST Letters and EMNIST Digits databases. For USPS, we train each model based on the first 3000 samples. For EMNIST Digits database, we randomly choose 10000 samples per class to form training set due to memory limit. The clustering curves under different numbers of the selected categories on USPS database are shown in Figs.8(e-g), and Tables 2-3 describe the averaged AC and F-scores according to the curves in Figs.8(e-g). Similar observations can be found from the results. That is, the AC and F-measure of each algorithm go down as the number of categories increases. It can also be found that SemiGNMF, CNMF and RS$^2$ACF deliver promising clustering results by using partial labeled data. Among the multilayer MF models, DSCF-Net obtains relatively better performance than

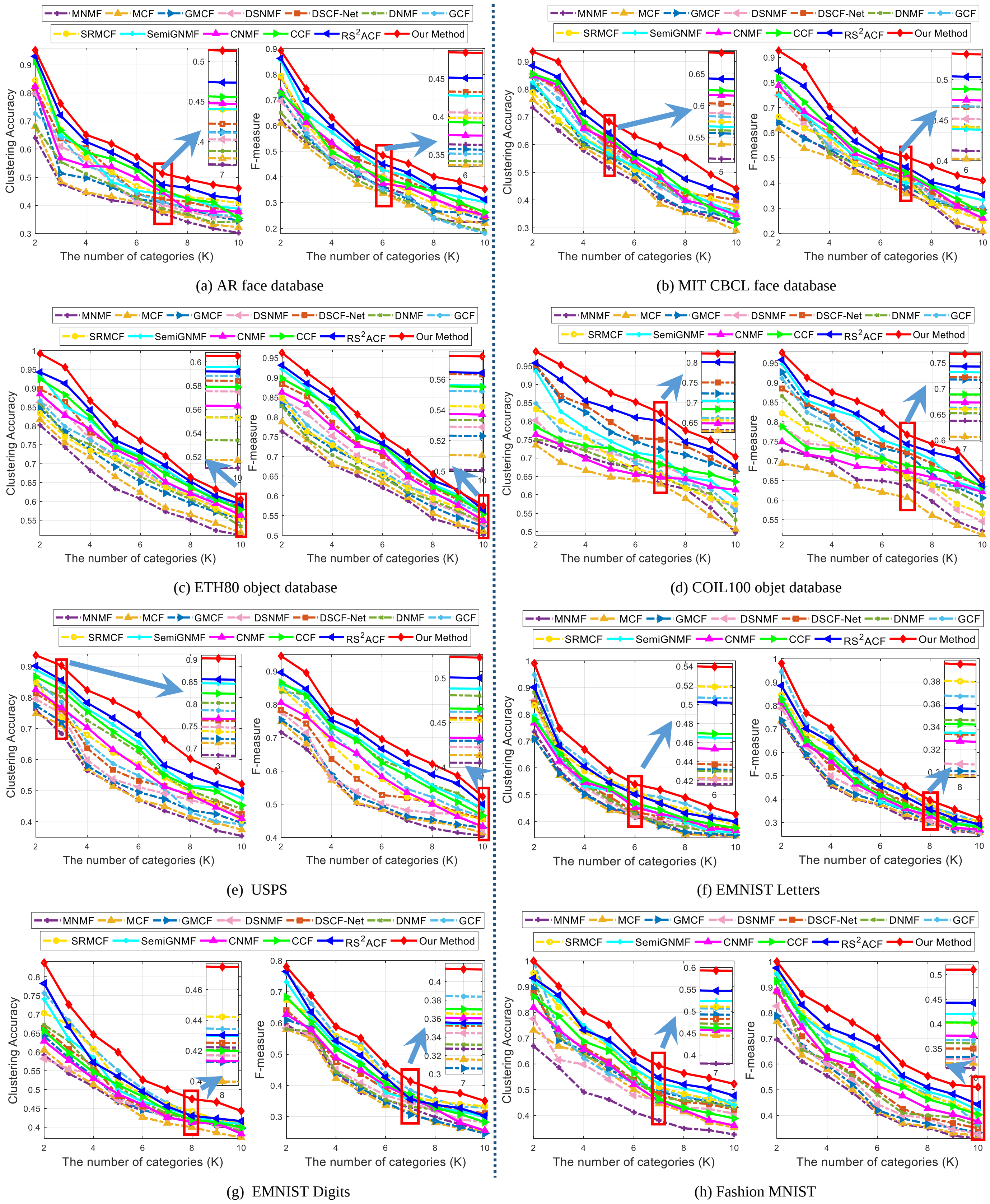

(a) AR face database (b) MIT CBCL face database

(c) ETH80 object database (d) COIL100 objet database

(e) USPS (f) EMNIST Letters

(g) EMNIST Digits (h) Fashion MNIST

**Figure 8:** Clustering performance over varied K values based on the eight evaluated databases.

**Table 2:** Averaged clustering accuracies (AC) of the algorithms based on the eight evaluated image databases.

| Methods | AR | | MIT CBCL | | ETH80 | | COIL100 | |
|---|---|---|---|---|---|---|---|---|
| | **Mean±std** | **Best** | **Mean±std** | **Best** | **Mean±std** | **Best** | **Mean±std** | **Best** |
| **MNMF** | 0.4127±0.1032 | 0.6406 | 0.4856±0.1461 | 0.7335 | 0.6251±0.1005 | 0.8021 | 0.6493±0.0801 | 0.7461 |
| **MCF** | 0.4277±0.1088 | 0.6807 | 0.4906±0.1651 | 0.7621 | 0.6430±0.1033 | 0.8176 | 0.6274±0.0710 | 0.7352 |
| **GMCF** | 0.4672±0.1367 | 0.7980 | 0.5179±0.1711 | 0.8078 | 0.6727± 0.1003 | 0.8462 | 0.7696± 0.0922 | 0.9492 |
| **DSNMF** | 0.4850±0.1483 | 0.7981 | 0.5640±0.1773 | 0.8550 | 0.6965±0.1055 | 0.8628 | 0.6855±0.0734 | 0.8019 |
| **DSCF-Net** | 0.5142±0.1476 | 0.8181 | 0.5721±0.1700 | 0.8491 | 0.7240±0.1094 | 0.8980 | 0.7846±0.0899 | 0.9516 |
| **DNMF** | 0.4539±0.1140 | 0.6820 | 0.5351±0.1595 | 0.8526 | 0.6790±0.1047 | 0.8596 | 0.6632±0.0712 | 0.7502 |
| **GCF** | 0.4882±0.1285 | 0.7265 | 0.5451±0.1802 | 0.8460 | 0.7003±0.0956 | 0.8663 | 0.6949±0.0904 | 0.8483 |
| **SRMCF** | 0.5300±0.1455 | 0.8449 | 0.5414±0.1389 | 0.7888 | 0.6807±0.0923 | 0.8337 | 0.6888±0.0913 | 0.8325 |
| **SemiGNMF** | 0.5289±0.1727 | 0.9231 | 0.5452±0.1575 | 0.8249 | 0.7335±0.1160 | 0.9355 | 0.7313±0.1061 | 0.9351 |
| **CNMF** | 0.5060±0.1404 | 0.8231 | 0.5650±0.1841 | 0.8479 | 0.7088±0.1109 | 0.8850 | 0.6721±0.0510 | 0.7640 |
| **CCF** | 0.5444±0.1693 | 0.9112 | 0.5733±0.1906 | 0.8550 | 0.7316±0.1228 | 0.9244 | 0.7044±0.0485 | 0.7842 |
| **RS$^2$ACF** | 0.5779±0.1648 | 0.9303 | 0.6134±0.1704 | 0.8848 | 0.7498±0.1266 | 0.9426 | 0.8126±0.0897 | 0.9578 |
| **Our method** | **0.6107±0.1610** | **0.9520** | **0.6654±0.1713** | **0.9355** | **0.7782±0.1386** | **0.9920** | **0.8479±0.0952** | **0.9889** |

| Methods | USPS | | EMNIST Letters | | EMNIST Digits | | Fashion MNIST | |
|---|---|---|---|---|---|---|---|---|
| | **Mean±std** | **Best** | **Mean±std** | **Best** | **Mean±std** | **Best** | **Mean±std** | **Best** |
| **MNMF** | 0.5077±0.1399 | 0.7601 | 0.4619±0.1304 | 0.7356 | 0.4676±0.0669 | 0.5915 | 0.4465±0.1183 | 0.6692 |
| **MCF** | 0.5173±0.1350 | 0.7479 | 0.4527±0.1225 | 0.7113 | 0.4595±0.0815 | 0.6050 | 0.5214±0.1377 | 0.7313 |
| **GMCF** | 0.5369±0.1324 | 0.7739 | 0.4583±0.1210 | 0.7086 | 0.4811±0.0871 | 0.6422 | 0.5770±0.1486 | 0.8935 |
| **DSNMF** | 0.5583±0.1315 | 0.7950 | 0.4929±0.1581 | 0.8522 | 0.4719±0.0703 | 0.5822 | 0.5254±0.1204 | 0.7756 |
| **DSCF-Net** | 0.5772±0.1346 | 0.8129 | 0.5070±0.1571 | 0.8416 | 0.4946±0.0943 | 0.6643 | 0.5784±0.1523 | 0.8759 |
| **DNMF** | 0.6247±0.1459 | 0.8400 | 0.4968±0.1726 | 0.8822 | 0.4982±0.1010 | 0.6722 | 0.5782±0.1609 | 0.9265 |
| **GCF** | 0.5734±0.1649 | 0.8530 | 0.5682±0.1705 | 0.9471 | 0.5367±0.1223 | 0.7567 | 0.5930±0.1759 | **1.000** |
| **SRMCF** | 0.5969±0.1425 | 0.8501 | 0.5419±0.1412 | 0.8343 | 0.5254±0.1103 | 0.7034 | 0.6268±0.1778 | 0.9531 |
| **SemiGNMF** | 0.6596±0.1524 | 0.8885 | 0.4995±0.1425 | 0.7963 | 0.5142±0.1149 | 0.7401 | 0.6361±0.1714 | 0.9220 |
| **CNMF** | 0.5954±0.1458 | 0.8252 | 0.4996±0.1318 | 0.7682 | 0.4796±0.0840 | 0.6291 | 0.5498±0.1622 | 0.8188 |
| **CCF** | 0.6425±0.1499 | 0.8668 | 0.5122±0.1346 | 0.7815 | 0.4967±0.0894 | 0.6566 | 0.5821±0.1739 | 0.8620 |
| **RS$^2$ACF** | 0.6780±0.1493 | 0.9016 | 0.5502±0.1611 | 0.9010 | 0.5306±0.1252 | 0.7822 | 0.6537±0.1643 | 0.9335 |
| **Our method** | **0.7276±0.1482** | **0.9355** | **0.6031±0.1776** | **0.9902** | **0.5800±0.1343** | **0.8375** | **0.7026±0.1707** | **1.000** |

**Table 3:** Averaged F-score values of the algorithms based on the eight evaluated real databases.

| Methods | AR | | MIT CBCL | | ETH80 | | COIL100 | |
|---|---|---|---|---|---|---|---|---|
| | **Mean±std** | **Best** | **Mean±std** | **Best** | **Mean±std** | **Best** | **Mean±std** | **Best** |
| **MNMF** | 0.3787±0.1423 | 0.6210 | 0.4132±0.1520 | 0.6388 | 0.6211±0.0906 | 0.7634 | 0.6379±0.0724 | 0.7277 |
| **MCF** | 0.3653±0.1345 | 0.6094 | 0.4033±0.1354 | 0.6150 | 0.6329±0.0950 | 0.7875 | 0.6124±0.0649 | 0.6935 |
| **GMCF** | 0.3967±0.1582 | 0.7117 | 0.4423±0.1246 | 0.6420 | 0.6559±0.1043 | 0.8322 | 0.7428±0.1038 | 0.9263 |
| **DSNMF** | 0.4205±0.1476 | 0.6956 | 0.4654±0.1596 | 0.7485 | 0.6764±0.1112 | 0.8452 | 0.6814±0.0819 | 0.7885 |
| **DSCF-Net** | 0.4559±0.1689 | 0.7881 | 0.4860±0.1580 | 0.7522 | 0.7091±0.1138 | 0.8835 | 0.7515±0.0848 | 0.8848 |
| **DNMF** | 0.3701±0.1561 | 0.6498 | 0.4705±0.1696 | 0.7882 | 0.6541±0.0942 | 0.8284 | 0.7072±0.0927 | 0.8655 |
| **GCF** | 0.3937±0.1959 | 0.7942 | 0.4887±0.1713 | 0.8173 | 0.6757±0.0986 | 0.8530 | 0.7270±0.0994 | 0.9109 |
| **SRMCF** | 0.4326±0.1777 | 0.7934 | 0.4490±0.1483 | 0.6633 | 0.6678±0.0996 | 0.8488 | 0.6918±0.0856 | 0.8214 |
| **SemiGNMF** | 0.4855±0.1801 | 0.8541 | 0.4925±0.1477 | 0.7496 | 0.7268±0.1251 | 0.9210 | 0.7672±0.1040 | 0.9478 |
| **CNMF** | 0.4344±0.1636 | 0.7338 | 0.4962±0.1748 | 0.7876 | 0.6976±0.1134 | 0.8629 | 0.6810±0.0394 | 0.7488 |
| **CCF** | 0.4412±0.1544 | 0.7269 | 0.5140±0.1789 | 0.8169 | 0.7232±0.1211 | 0.8980 | 0.7037±0.0442 | 0.7875 |
| **RS$^2$ACF** | 0.5061±0.1816 | 0.8617 | 0.5509±0.1781 | 0.8460 | 0.7408±0.1269 | 0.9310 | 0.7874±0.0972 | 0.9578 |
| **Our method** | **0.5414±0.1817** | **0.8923** | **0.6047±0.1879** | **0.9262** | **0.7612±0.1355** | **0.9626** | **0.8144±0.1010** | **0.9762** |

| Methods | USPS | | EMNIST Letters | | EMNIST Digits | | Fashion MNIST | |
|---|---|---|---|---|---|---|---|---|
| | **Mean±std** | **Best** | **Mean±std** | **Best** | **Mean±std** | **Best** | **Mean±std** | **Best** |
| **MNMF** | 0.5165±0.1136 | 0.7162 | 0.4075±0.1572 | 0.7250 | 0.3853±0.1199 | 0.5944 | 0.4564±0.1384 | 0.6960 |
| **MCF** | 0.5272±0.1171 | 0.7479 | 0.4135±0.1587 | 0.7341 | 0.3761±0.1206 | 0.5795 | 0.4750±0.1562 | 0.7658 |
| **GMCF** | 0.5386±0.1186 | 0.7539 | 0.4212±0.1585 | 0.7364 | 0.3865±0.1346 | 0.6101 | 0.4930±0.1613 | 0.7878 |
| **DSNMF** | 0.5510±0.1201 | 0.7774 | 0.4404±0.1742 | 0.7877 | 0.4012±0.1116 | 0.5835 | 0.5044±0.1673 | 0.8270 |
| **DSCF-Net** | 0.5799±0.1170 | 0.7829 | 0.4699±0.1766 | 0.8182 | 0.4252±0.1175 | 0.6402 | 0.5354±0.1792 | 0.8833 |
| **DNMF** | 0.6560±0.1281 | 0.8370 | 0.4743±0.1785 | 0.8054 | 0.4022±0.1081 | 0.5801 | 0.5357±0.1873 | 0.9490 |
| **GCF** | 0.6268±0.1396 | 0.8551 | 0.5313±0.2106 | 0.9446 | 0.4835±0.1547 | 0.7593 | 0.6013±0.1802 | 0.9525 |
| **SRMCF** | 0.6050±0.1393 | 0.8501 | 0.5089±0.1803 | 0.8460 | 0.4650±0.1315 | 0.6736 | 0.6233±0.1841 | 0.9662 |
| **SemiGNMF** | 0.6607±0.1332 | 0.8600 | 0.4715±0.1874 | 0.8348 | 0.4478±0.1516 | 0.7315 | 0.6305±0.1719 | 0.9538 |
| **CNMF** | 0.6034±0.1335 | 0.8055 | 0.4572±0.1827 | 0.8150 | 0.4181±0.1290 | 0.6273 | 0.5606±0.1695 | 0.8856 |
| **CCF** | 0.6489±0.1399 | 0.8662 | 0.4709±0.1830 | 0.8263 | 0.4404±0.1383 | 0.6835 | 0.5939±0.1721 | 0.9238 |
| **RS$^2$ACF** | 0.6841±0.1319 | 0.8962 | 0.5093±0.1990 | 0.8848 | 0.4652±0.1585 | 0.7650 | 0.6570±0.1766 | 0.9752 |
| **Our method** | **0.7160±0.1401** | **0.9455** | **0.5625±0.2190** | **0.9806** | **0.5111±0.1512** | **0.7800** | **0.7045±0.1721** | **1.000** |

other methods. By further enriching the supervised prior by predicting the labels of unlabeled data, and designing a more reasonable dual-constrained deep structures, our DS$^2$CF-Net outperforms all its competitors by delivering better results.

**Fashion Products Clustering.** Finally, we test each method for representing the fashion product images of Fashion MNIST database. We train each model by a subset of Fashion MNIST, i.e., totally 10,000 samples from 10 classes. The clustering results in terms of AC and F-measure are evaluated and shown in Fig.8(h). Tables 2-3 describe the statistics in terms of averaged AC and F-scores according to Fig.8(h). From the results, we can similarly see that: 1) our DS$^2$CF-Net delivers enhanced performance than other competitors in most cases, especially when the number of K is relatively small. We also find that semi-supervised methods can generally deliver enhanced performance than unsupervised ones. But note that SRMCF also obtains the promising results, which implies that the self-expression property is also important to improve the representation ability. It should be noted that our DS$^2$CF-Net also employs the self-expression scheme in the proposed multilayer structures. In addition, one can also find that the results of the multilayer MNMF, MCF and GMCF models are worse than those of the single-layer models, which verifies that their multilayer structures of directly feeding the learnt representation from the last layer to the next layer is indeed not reasonable.

### *D. Ablation Study*

**(1) Clustering with different proportions of labeled samples**. We first evaluate each semi-supervised factorization model, i.e., CNMF, CCF, SemiGNMF, RS$^2$ACF and our DS$^2$CF-Net, by using different numbers of labeled data in each class. In this study, for each database the proportion of labeled samples varies from 10% to 90%, and we randomly choose three categories for this test. We average the results over 10 random selections of categories and 30 initializations for the K-means clustering for each MF approach to avoid the randomness. For comparison, we also report the clustering results of four representative unsupervised methods (i.e., MCF, DSNMF, GMCF, and DSCF-Net) as baselines using the flat dashed lines. The clustering results based on the evaluated databases are reported in Fig.9. We see that: (1) the increasing number of labeled samples can greatly improve the clustering performance of each method. It can also be found that the improvement by our DS$^2$CF-Net over other compared methods is more obvious, especially when the proportion of label data is relatively small; (2) our DS$^2$CF-Net delivers better results across different labeled proportions by fully mining the intrinsic relations between the labeled and unlabeled data, and predicting the labels of unlabeled samples to enrich the supervised prior knowledge. RS$^2$ACF also performs well by delivering better results than other remaining methods.

**Further discussion between semi-supervised and unsupervised CF-based clustering.** As shown in Fig.9, unsupervised methods (such as DSCF-Net) obtain better results than DS$^2$CF-Net when the proportion of labeled samples is low (such as 10%), especially on the AR, ETH80, COIL100 and USPS datasets. The major reasons are twofold. First, although DSCF-Net is an unsupervised method and cannot preserve the locality in feature space compared with DS$^2$CF-Net, it clearly incorporates the noise removal process in its model. That is, DSCF-Net performs the factorization in the noise-removed clean data space, while DS$^2$CF-Net performs in the original space; Second, even though the formulation of our DS$^2$CF-Net looks like that of DSCF-Net when DS$^2$CF-Net does not exploit labeled data, note that there are several obvious differences between them: 1) although they both are deep matrix factorization methods, their factorization mechanisms are totally different. Specifically, DSCF-Net uses a single-channel mode, which optimizes the basis vectors to update the representation matrix indirectly in each layer. While DS$^2$CF-Net designs a two-channel factorization model, which can jointly update the basis vectors and representation matrix in each layer; 2) DSCF-Net can only retain the local information of the data space, while our DS$^2$CF-Net can preserve the local manifold structures of both the data space and feature space. Therefore, DSCF-Net and DS$^2$CF-Net are totally different two methods. In other words, DS$^2$CF-Net is not a simple extension of DSCF-Net. As a result, DSCF-Net has a potential to outperform DS$^2$CF-Net in some cases, for example when the proportion of labeled samples is relatively low.

**(2) Clustering with different numbers of layers.** We investigate the effects of the number of layers on the representation learning and clustering abilities of each multilayer model, including MNMF, MCF, GMCF, DSNMF, DSCF-Net and our DS$^2$CF-Net. In this simulation, we vary the number of layers from 1 to 10 with step 1. For each database, we randomly choose 3 categories for the clustering evaluations. The averaged clustering ACs are illustrated in Fig.10, from which we see that: 1) our DS$^2$CF-Net delivers the highest accuracies than other methods in most cases; 2) the increase of the number of layers can generally improve the clustering results, which implies that discovering hidden deep features can indeed improve the performance. However, the clustering results of MNMF, DSNMF, MCF and GMCF go down apparently when the number of layers passes 4 in most cases, which maybe because MNMF, MCF and GMCF cannot ensure the intermediate representation from the previous layer to be a good representation for subsequent layers. Note that the first stage of DSNMF is performed similarly as MNMF, MCF and GMCF, thus it also suffers from the degrading issue as the number of layers is increased to a high level. This observation can once again show that the multilayer structures of directly feeding the learnt representation from the last layer to the next layer is not reasonable. By updating the basis vectors to optimize the low-dimensional representation in each layer, DSCF-Net also perform well by delivering higher and more stable results than the other remaining methods, i.e., MNMF, DSNMF, MCF and GMCF. By this analysis, we can choose a proper number of layers for each multilayer MF model for the representation and clustering tasks in the experiments.

**(3) Hyperparameter sensitivity analysis.** We investigate the effects of the hyper-parameters of each compared method on the representation and clustering abilities. For the compared methods, five methods (MNMF, MCF, DSNMF, CNMF and CCF) have no hyper-parameter in their models; two methods (GMCF and SemiGNMF) have one hyper-parameter $\alpha$ in their models; three methods (DNMF, GCF and SRMCF) contain two hyper-parameters $\alpha$ and $\beta$ in their models; last two algorithms (DSCF-Net and RS$^2$ACF), have three hyper-parameters $\alpha$, $\beta$ and $\gamma$ in their models. Note that we uniformly use $\alpha$, $\beta$ and $\gamma$ to represent the first, the second and the third hyper-parameters of each method if have, and all hyper-parameters will be selected from the same candidate set {$10^{-5}$, $10^{-4}$,…,$10^5$} for fair comparison. Specifically, for the method with only one parameter $\alpha$, we directly tune it from the candidate set and evaluate the performance. For the methods with two parameters $\alpha$ and $\beta$, we adopt the commonly-used grid search strategy [32-33] to tune them from the candidate set. For the methods with three hyper-parameters, we first fix $\gamma=1$ and tune $\alpha$ and $\beta$ using the grid search strategy, and then fix the selected $\alpha$ and $\beta$ to tune $\gamma$. For RS$^2$ACF, a fixed parameter setting, i.e., $\alpha=10^4$, $\beta=10^{-4}$ and $\gamma=10^4$, was provided in [42], so we directly use this setting in our experiments. For each database, we choose the samples of three categories to train each method, set the number of layers is set to 3 for each multi-layer method and the proportion of labeled data is still set to 40% for each semi-supervised method. The results are averaged over 30 random initializations of the cluster centers for the K-means clustering algorithm. The hyper-parameters sensitivity analysis results of each method over AR and COIL100 databases are displayed in Figs.11-13 as examples. Finally, we report the best choice of the hyperparameters of each method in Table 4, which have also been used in the clustering evaluations of this paper.

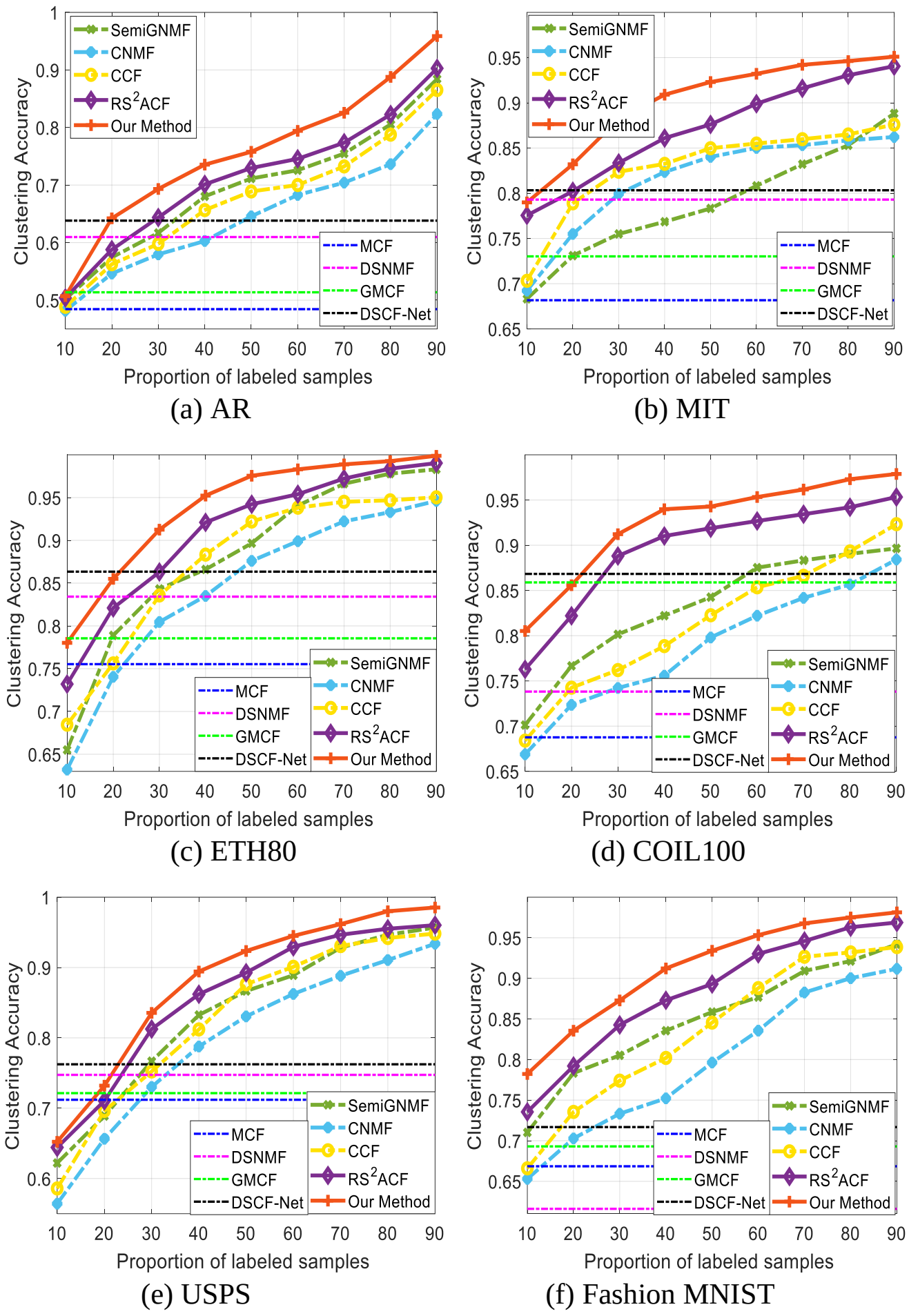

**Figure 9:** Clustering accuracies vs. varied proportions of labeled samples based on the evaluated image databases.

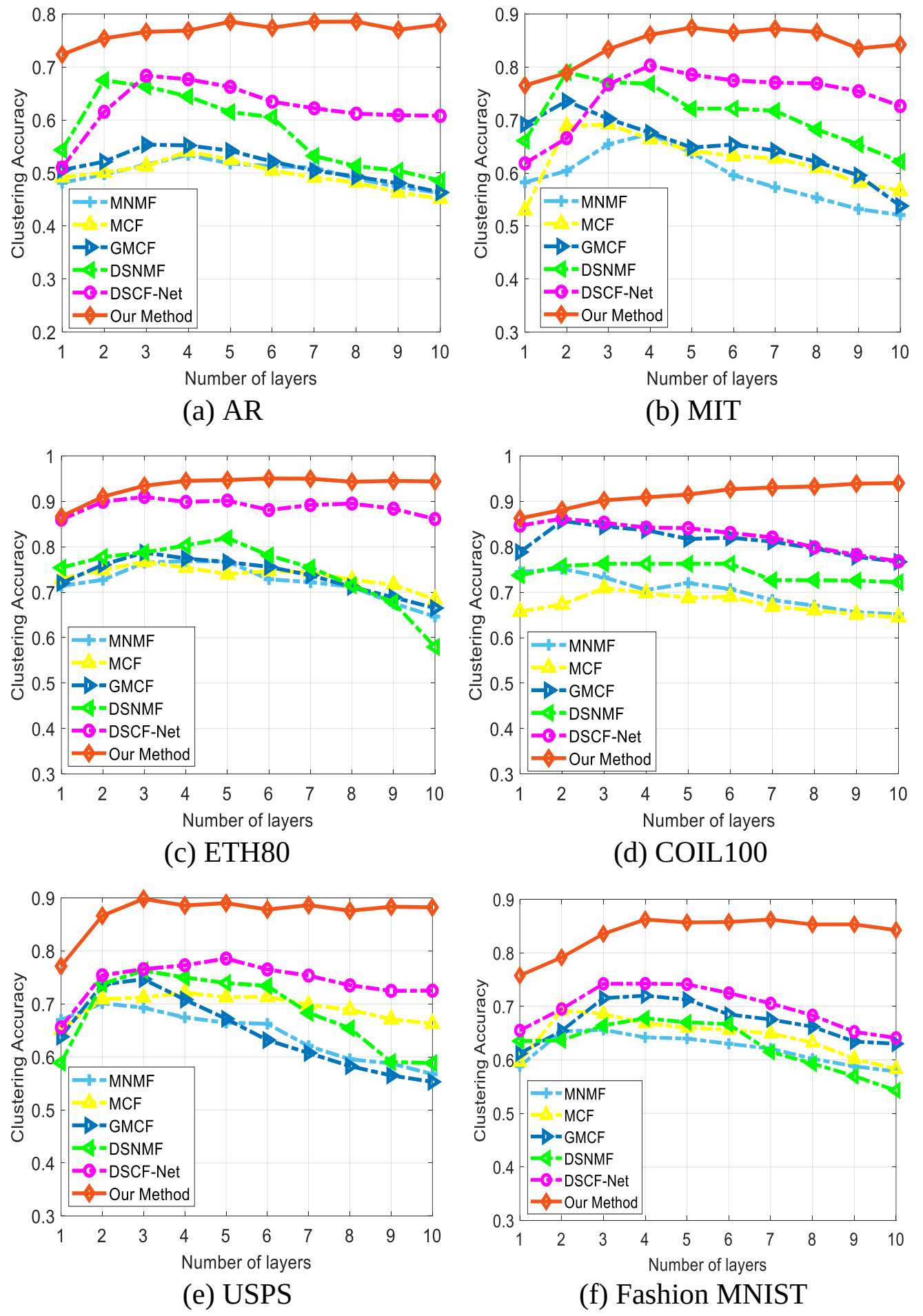

**Figure 10:** Clustering accuracies vs. varied number of layers based on the evaluated image databases.

**Table 4.** Settings of parameters for each algorihtm based on the evaluated image databases.

| Method | AR | MIT CBCL | ETH80 | COIL100 | USPS | EMNIST Letters | EMNIST Digits | Fashion MNIST |
|---|---|---|---|---|---|---|---|---|
| GMCF (α) | 1 | 10 | 1 | 10 | 1 | 1 | 1 | 0.1 |
| DNMF (α,β) | $10^{-4}$, 1 | $10^5$, $10^{-4}$ | 10, 10 | 1, 1 | 1, 10 | $10^{-4}$, 1 | $10^{-4}$, 1 | 1, 10 |
| GCF (α,β) | $10^3$, 1 | $10^5$, $10^5$ | 1, $10^2$ | $10^5$, $10^{-5}$ | $10^5$, $10^{-5}$ | $10^3$, 1 | $10^3$, 1 | $10^5$, $10^2$ |
| SRMCF (α,β) | $10^{-3}$, $10^3$ | $10^{-5}$, $10^5$ | $10^{-5}$, $10^5$ | $10^{-5}$, $10^5$ | $10^{-5}$, $10^5$ | $10^{-3}$, $10^3$ | $10^{-3}$, $10^3$ | $10^{-5}$, $10^5$ |
| SemiGNMF (α) | 10 | 10 | 1 | $10^2$ | 10 | 10 | 10 | 1 |
| DSCF-Net (α,β,γ) | $10^{-4}$, 1, $10^3$ | $10^{-5}$, 10, $10^2$ | $10^{-3}$, 10, $10^3$ | 10, $10^{-1}$, $10^{-2}$ | $10^{-1}$, 1, $10^3$ | $10^{-4}$, 10, $10^3$ | $10^{-4}$, 10, $10^3$ | $10^{-2}$, 10, 10 |
| DS$^2$CF-Net (α,β,γ) | $10^{-4}$, $10^2$, $10^{-4}$ | 1, $10^2$, $10^{-3}$ | 1, 1, $10^{-2}$ | $10^{-3}$, 10, $10^{-1}$ | $10^{-1}$, 1, $10^{-2}$ | $10^{-2}$, $10^{-2}$, $10^{-2}$ | $10^{-2}$, $10^{-2}$, $10^{-2}$ | $10^{-2}$, 10, 1 |

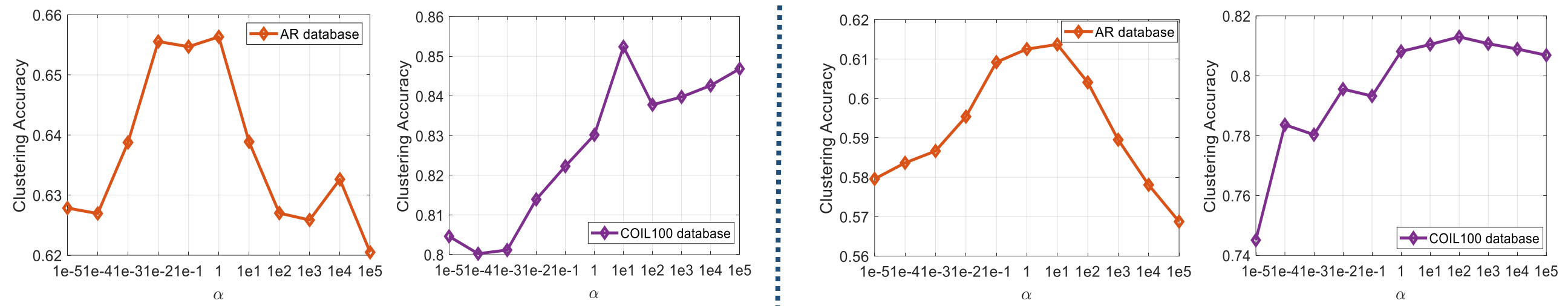

(a) GMCF (b) SemiGNMF

**Figure 11:** Clustering accuracies of GMCF and SemiGNMF under various parameters over AR database and COIL100 database.

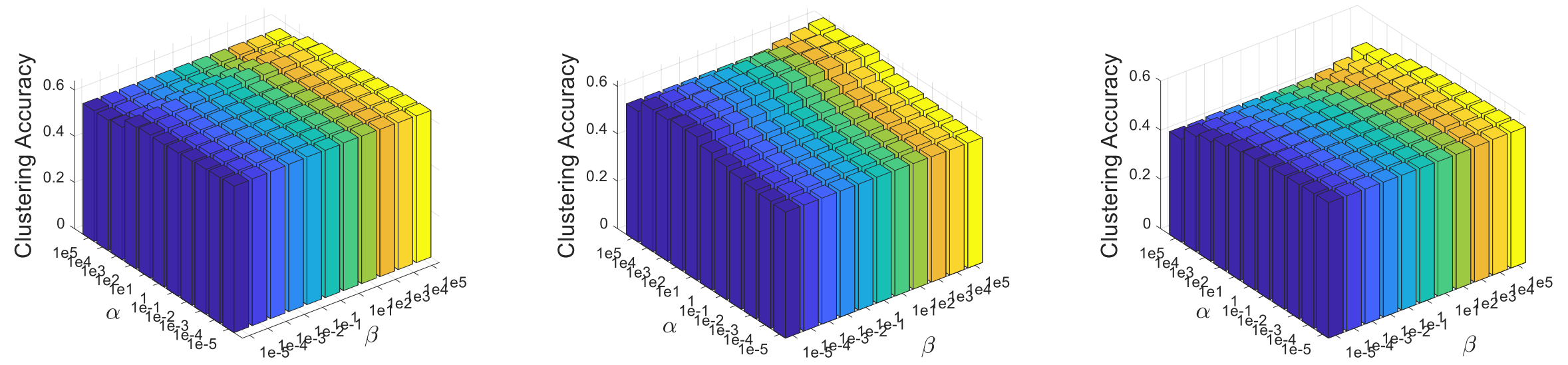

(a) DNMF on AR database (b) GCF on AR database (c) SRMCF on AR database

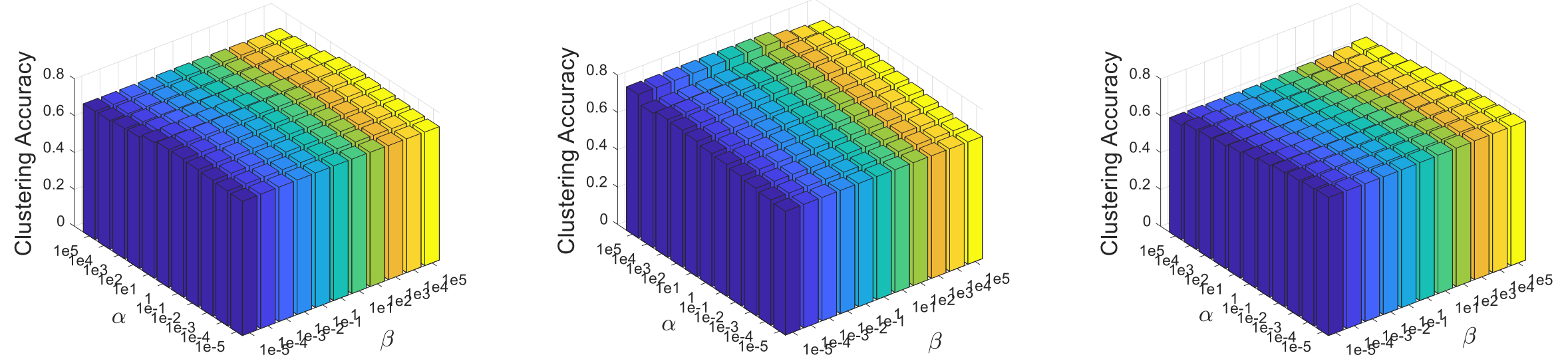

(d) DNMF on COIL100 database (e) GCF on COIL100 database (f) SRMCF on COIL100 database

**Figure 12:** Clustering accuracies of DNMF, GCF and SRMCF under various parameters over AR database and COIL100 database.

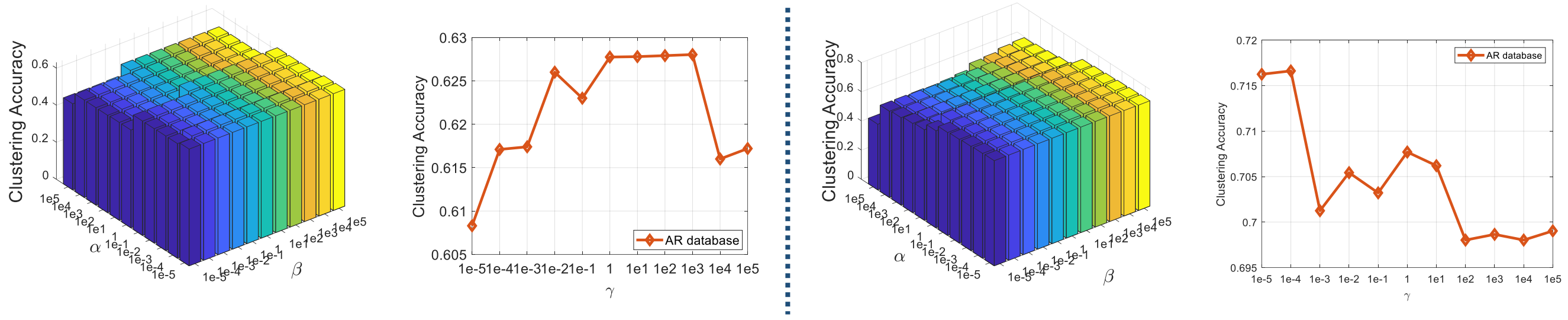

(a) DSCF-Net on AR database (b) DS$^2$CF-Net on AR database

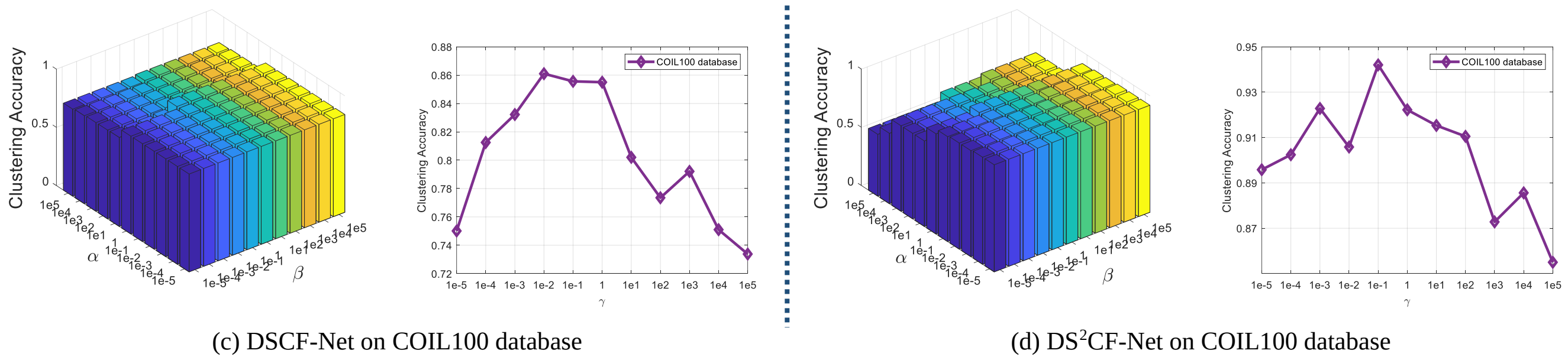

(c) DSCF-Net on COIL100 database (d) DS$^2$CF-Net on COIL100 database

**Figure 13:** Clustering accuracies of DSCF-Net and DS$^2$CF-Net under various parameters over AR database and COIL100 database.

## VI. Concluding Remarks

We proposed a new enriched prior knowledge guided dual-constrained deep semi-supervised coupled factorization model for discovering hierarchical information. To capture hidden deep information, our DS$^2$CF-Net designs a joint label and structure-constrained factorization network using multiple layers of linear transformations of basis vectors and representations. An error correction mechanism with a feature fusion strategy is also integrated between consecutive layers to improve the representation. To improve the discrimination of deep representation and coefficients, DS$^2$CF-Net clearly considers enriching the supervised prior knowledge by the joint deep coefficients-regularized label prediction, and incorporates enriched prior information as additional label and structure constraints. Moreover, DS$^2$CF-Net also proposes to keep the locality structures in both the data and feature spaces by adopting an adaptive dual-graph weighting strategy. A fine-tuning process is finally included to refine the structure-constrained matrix and the data weight matrix in each layer using the predicted labels for more accurate representations.

We have evaluated our DS$^2$CF-Net for image representation and clustering, and the results are compared with several related single-layer and multilayer frameworks. Both the visual image analysis and quantitative clustering evaluation demonstrate the effectiveness of our framework. In future, we will evaluate our method for the other related application areas, such as document retrieval and recommended system. More efficient coupled factorization strategy will also be investigated for the consideration of scalability. In addition, we will explore how to integrate the factorization model with the deep convolutional neural network for handling the large-scale vision tasks.

## Acknowledgments

The authors also would like to express our sincere thanks to the anonymous reviewers' constructive comments and suggestions which have made the paper a higher standard. We also sincerely thank Prof. Mingliang Xu and Prof. Yi Yang for their professional discussion on the error correction mechanism with feature fusion strategy, and the fine-tuning process for refining the features. This work is partially supported by the National Natural Science Foundation of China (62072151, 62020106007, 61806035 and U1936217), Anhui Provincial Natural Science Fund for Distinguished Young Scholars (2008085J30) and the Fundamental Research Funds for the Central Universities of China (JZ2019-HGPA0102). Zhao Zhang is the corresponding author of this present paper.